%% file: acl_latex.tex
\pdfoutput=1

\documentclass[11pt]{article}

\usepackage[preprint]{acl}
\usepackage{times}
\usepackage{multicol}
\usepackage{enumitem}
\usepackage{latexsym}
\usepackage{booktabs}
\usepackage{amssymb}
\usepackage{amsmath}
\usepackage{bbding}
\usepackage{subfig}
\usepackage{makecell}
\usepackage{wrapfig}
\usepackage{graphicx}
\usepackage{multirow}
\usepackage[ruled,linesnumbered]{algorithm2e}
\usepackage{colortbl}
\usepackage{rotating}
\usepackage[inkscapelatex=false]{svg}
\usepackage{url}
\usepackage{diagbox}
\usepackage{pifont}
\usepackage{listings}
\usepackage{orcidlink}

\usepackage[T1]{fontenc}
\usepackage[utf8]{inputenc}
\usepackage{microtype}
\usepackage{inconsolata}
\newcommand{\tikzhl}[2]{\tikz[baseline=(x.base)]{\node(x)[rectangle, fill={#1}, rounded corners, text height=4.5pt]{#2};}}
\newcommand{\tikzhlforeq}[2]{\tikz[baseline=(x.base)]{\node(x)[rectangle, fill={#1}, rounded corners]{#2};}}
\definecolor{punishred}{HTML}{ffcccc}
\definecolor{rewardblue}{HTML}{cce5ff}
\renewcommand{\paragraph}{\vspace{0.3em}\noindent\textbf}
\newcommand{\std}[1]{\textcolor{gray}{${}_{#1}$}}

\title{Mechanistic Fine-tuning for In-context Learning}

\author{
Hakaze Cho\orcidlink{0000-0002-7127-1954}${}^{1,\text{\ding{73}}}$\phantom{1.}
Peng Luo${}^{2}$\phantom{1.}
Mariko Kato${}^{1}$\phantom{1.}
Rin Kaenbyou${}^{2}$\phantom{1.}
Naoya Inoue${}^{1,3}$\\
${}^{1}$Japan Advanced Institute of Science and Technology\\${}^{2}$Beijing Institute of Technology\phantom{11}${}^{3}$RIKEN\\
${}^{\text{\ding{73}}}$\hyperref[Appendix:ACS]{Primary Contributor}, Correspondence to: \texttt{yfzhao@jaist.ac.jp}
}

\begin{document}
\maketitle
\begin{abstract}
\textbf{I}n-\textbf{c}ontext \textbf{L}earning (ICL) utilizes structured demonstration-query inputs to induce few-shot learning on \textbf{L}anguage \textbf{M}odels (LMs), which are not originally pre-trained on ICL-style data. To bridge the gap between ICL and pre-training, some approaches fine-tune LMs on large ICL-style datasets by an end-to-end paradigm with massive computational costs. To reduce such costs, in this paper, we propose \textbf{A}ttention \textbf{B}ehavior \textbf{F}ine-\textbf{T}uning (ABFT), utilizing the previous findings on the inner mechanism of ICL, building training objectives on the attention scores instead of the final outputs, to force the attention scores to focus on the correct label tokens presented in the context and mitigate attention scores from the wrong label tokens. Our experiments on 9 modern LMs and 8 datasets empirically find that ABFT outperforms in performance, robustness, unbiasedness, and efficiency, with only around 0.01\% data cost compared to the previous methods. Moreover, our subsequent analysis finds that the end-to-end training objective contains the ABFT objective, suggesting the implicit bias of ICL-style data to the emergence of induction heads. Our work demonstrates the possibility of controlling specific module sequences within LMs to improve their behavior, opening up the future application of mechanistic interpretability\footnote{Source code of this paper is available at~\url{https://github.com/hc495/ICL_head_tuning}.}.
\end{abstract}

\section{Introduction}

\begin{figure*}
    \centering
    \includegraphics[width=0.95\linewidth]{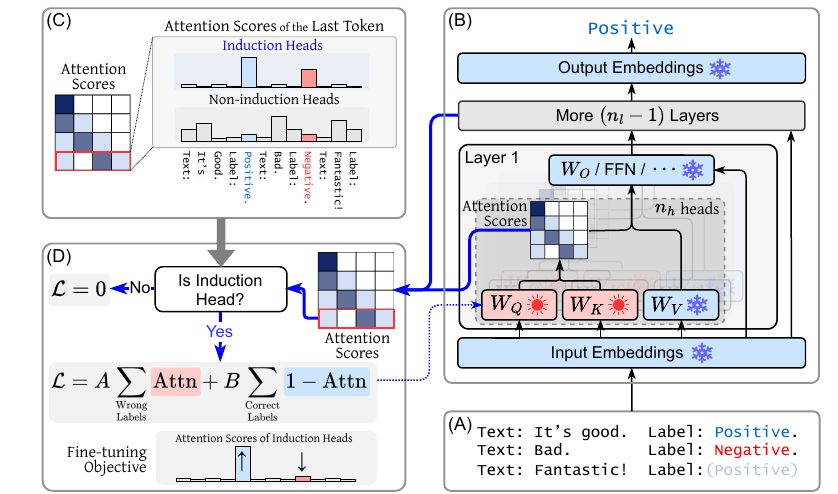}
    \vspace{-0.5\baselineskip}
    \caption{Diagram of ABFT framework. (A) \textbf{An example of ICL-style inputs.} We build datasets from such examples to fine-tune models. (B) \textbf{Feed-forward inference of ICL.} We collect the attention scores of every attention head in every layer to calculate the training objective. and we only enable the gradient of the $W_Q$ and $W_K$ matrices. (C) \textbf{The criterion for induction head.} Only attention heads producing attention scores with a significant focus on the label tokens can be identified as induction heads. (D) \textbf{Loss calculation of ABFT.} Only induction heads return a non-zero loss, and such loss contains a punishment on ``wrong'' attention scores to wrong label tokens, and a reward on ``correct'' attention scores to correct label tokens.}
    \label{fig:main_figure}
    \vspace{-\baselineskip}
\end{figure*}

\textbf{I}n-\textbf{C}ontext \textbf{L}earning (ICL)~\citep{radford2019language, dong2022survey} is an emerging few-shot learning paradigm where only a concatenation of few-shot \textit{demonstrations} and a \textit{query} is needed to conduct the specified task on the query, requiring only feed-forward calculation on the pre-trained \textbf{L}anguage \textbf{M}odels (LMs), as shown in Fig.~\ref{fig:main_figure} (A, B). However, trained on natural language data, LMs may face a distribution gap with ICL-style inputs, potentially hindering ICL performance. Therefore, some prior studies (see~\S\ref{sec:background}) try to bridge such a gap by fine-tuning LMs on the ICL-style data on end-to-end paradigms, with enormous datasets and calculation cost, preventing practical application, especially on the scaling \textbf{L}arge \textbf{LM}s (LLMs).

Therefore, in this paper, we try to propose an efficient fine-tuning approach towards better ICL performance, utilizing some previous observations on the inner mechanisms of ICL. In detail, we focus on the \textit{Induction Heads} in Transformer-based LMs, which are a set of critical attention heads towards ICL, where the attention scores of the last token in the ICL input (where the predictions are generated) are dominant on the label tokens in the demonstrations, as shown in Fig.~\ref{fig:main_figure} (C), for the clue that the tendency of attention scores from induction heads influences the tendency of prediction synchronously~\cite{reddy2023mechanistic,cho2025revisiting} (e.g., if the attention scores of the induction heads focus on the label token ``negative'' in the context, then the prediction is biased towards ``negative''). 

Consequently, we can directly control the attention scores to make the induction heads focus on the correct label tokens for correct predictions. Given such an objective, as shown in Fig.~\ref{fig:main_figure} (C, D), we propose \textbf{A}ttention \textbf{B}ehavior \textbf{F}ine-\textbf{T}uning (ABFT), \textbf{calculating fine-tuning objective (loss function) only on the attention scores of induction heads}, to mitigate ``wrong'' attention score focusing on the wrong label tokens, and promote ``correct'' attention score focusing on the correct label tokens. On such an objective, we fine-tune only the $W_K$ and $W_Q$ projection matrices of every attention head, with an ICL-style training set of only a few hundred samples, and only a few million of the parameters with gradient activated, which is highly efficient compared to previous works.

Our experiments on 9 modern (L)LMs and 8 downstream datasets demonstrate that ABFT significantly improves ICL performance with satisfactory efficiency, robustness, unbiasedness, and harmlessness, even outperforming previous works of end-to-end fine-tuning the whole model on massive datasets that are approximately $7,000\times$ larger than ours. Moreover, our analysis shows that the ABFT objective implicitly biases end-to-end training on ICL-style data, where causal language modeling may naturally induce induction heads.


\paragraph{Our contribution can be summarized as:}
\begin{itemize}[topsep=0pt, itemsep=-3pt, leftmargin=12pt]
    \item We propose \textbf{A}ttention \textbf{B}ehavior \textbf{F}ine-\textbf{T}uning (ABFT), which efficiently fine-tunes LMs on ICL inputs using attention-based objectives without supervision on the final output.
    \item Subsequent analysis indicates that the training objective of ABFT is implicitly encompassed by the end-to-end training objective on ICL-style data, suggesting that these data may naturally evoke the induction heads, which enhances the previous works on the inner mechanism of ICL.
    \item Also, we prototypically confirm the possibility of optimizing model performance directly by controlling the intermediate behavior, without any error propagation from the output. This is a hint toward \textit{Mechanistic Controllability}, a valuable future of mechanistic interpretability.
\end{itemize}

\section{Background \& Related Works}
\label{sec:background}

\paragraph{In-context Learning.} Given a few-shot \textit{demonstration} set $\{(x_i, y_i)\}_{i=1}^k$ and a query $x_q$, typical ICL creates a concatenation formed like $[x_1,y_1,x_2,y_2,\dots,x_k,y_k,x_q]$, and feeds it into the forward calculation of a pre-trained LM~\cite{radford2019language, dong2022survey} for the next token as the prediction to $x_q$, as shown in Fig.~\ref{fig:main_figure} (A).

\paragraph{LM Warm-up for ICL.} Since LMs are typically pre-trained on plain natural language data instead of ICL-style data, it can be expected that a distribution gap between the pre-training and ICL testing occurs to prevent optimal performance. Therefore, some works focus on tuning LMs on the ICL data~\cite{chen2022improving, min2022metaicl, mishra2022cross, wang2022super, wei2023symbol}. Even effective, these works need gradient-based whole-model and full-precision training on large datasets, making it hard to adapt to real-world applications due to the calculation overhead, and misaligning with the low-resource purpose of ICL. 

\paragraph{Induction Heads in ICL Inference.} As shown in Fig.~\ref{fig:main_figure} (C), it has been found that some attention heads (called \textit{induction heads}) in LMs have a nontrivial influence on ICL inference~\cite{elhage2021mathematical, olsson2022context, singh2024needs, reddy2023mechanistic, cho2025revisiting}, where the attention scores from the last token of the query (the location for prediction, e.g., the last ``: '' in Fig.~\ref{fig:main_figure} (A), as the attention query) concentrate on the label tokens presented in the demonstrations (e.g., ``positive''s and ``negative''s in Fig.~\ref{fig:main_figure} (A), as the attention key). Attention connections from induction heads transfer label information from the demonstration to the output, biasing predictions toward labels with higher attention scores. Consequently, the accuracy of ICL prediction critically depends on whether these attentions are on the correct labels.

\section{Attention Behavior Fine-tuning}
\label{sec:method}

Given the inspiration from the previous works, where the ICL predictions are biased towards the more attention-score-concentrated labels in the induction heads, in this paper, as shown in Fig.~\ref{fig:main_figure}, we propose \textbf{A}ttention \textbf{B}ehavior \textbf{F}ine-\textbf{T}uning (ABFT), a novel low-resource fine-tuning method to induce attention scores to focus on the correct labels. 

\paragraph{Method Pipeline.} Globally, ABFT utilizes such a pipeline: \textbf{(1) Dataset Building}: from a selected downstream dataset, we build a training set composed of ICL-style sequences as shown in Fig.~\ref{fig:main_figure} (A). \textbf{(2) Feed-forward Calculation}: For each training sample, as shown in Fig.~\ref{fig:main_figure} (B), we conduct a standard feed-forward calculation on the pre-trained LM, and collect the attention matrices of all the attention heads in all the layers. \textbf{(3) Loss Calculation}: For each attention matrix, we only focus on the attention scores of the last token (i.e., the last row of the attention matrix), where the predictions of queries are made. As shown in Fig.~\ref{fig:main_figure} (D), we first filter (detailed below) the non-induction head out, and return a loss of $0$ for these heads. Then, for the remaining induction heads, we calculate a loss function composed of a \tikzhl{punishred!80}{punishment} of attention scores on wrong labels and a \tikzhl{rewardblue!80}{reward} of attention scores on correct labels (detailed below). \textbf{(4) Back Propagation}: We back-propagate the calculated loss only to the $W_Q$ and $W_K$ matrices of every attention head, and update the model parameters.

\paragraph{Induction Head Filter.} As shown in Fig.~\ref{fig:main_figure} (C), we skip the attention matrices where the attention scores of the last token do not dominate on the label tokens. To identify the attention matrices to skip, in detail, given an attention matrix $\mathcal{A}\in\mathbb{R}^{n_t\times n_t}$, where the $n_t$ is the input token sequence length, as mentioned before, we focus on the last row $\mathbf{\alpha} = \mathcal{A}_{n_t}$. Given the position index of label tokens as $\mathcal{I}=\left\{\mathcal{I}_{i}\right\}_{i=1}^k$, we calculate the attention score sum on these label tokens as $S=\sum_{j\in\mathcal{I}} \mathbf{\alpha}_{j}$.\footnote{For the case shown in Fig.~\ref{fig:main_figure} (C), $\mathcal{I}=\{4, 8\}$ (0-started), and $S$ is the sum of the values of the red-highlighted bar and the blue-highlighted bar.} Then, we set a threshold $T=\frac{k}{k+\log(n_t)}$, if $S>T$, we assert the attention head of score $\mathcal{A}$ is an induction head, and vice versa. We will verify the necessity and benefits of this induction head filter in \S\ref{sec:ablations}.

\input{Table/Tab1_mainres}

\paragraph{Loss Function.} As shown in Fig.~\ref{fig:main_figure} (D), given an attention matrix $\mathcal{A}$, if judged as a non-induction head by the aforementioned head filter, the loss $\mathcal{L}$ for $\mathcal{A}$ is assigned to $0$. Else, we conduct the following calculation: given the position index of label tokens consistent with the ground-truth label of the query as $\mathcal{I}^+$, and the others $\mathcal{I}^-=\mathcal{I}\textbackslash\mathcal{I}^+$,\footnote{For the case shown in Fig.~\ref{fig:main_figure} (C), $\mathcal{I}^+=\{4\}$ (the position of ``positive''), $\mathcal{I}^-=\{8\}$ (the position of ``negative'').} we calculate the loss from the last row ($\mathbf{\alpha}$) of $\mathcal{A}$ as:
\vspace{-1\baselineskip}
\begin{equation}
    \label{eq:ABFTLoss}
    \mathcal{L}(\mathcal{A}) = A\enspace\text{\tikzhlforeq{punishred!80}{$\sum_{i\in\mathcal{I}^-}\mathbf{\alpha}_i$}} + B\enspace\text{\tikzhlforeq{rewardblue!80}{$\sum_{i\in\mathcal{I}^+}1-\mathbf{\alpha}_i$}}.
\end{equation}
\noindent That is, as shown in Fig.~\ref{fig:main_figure} (D), we \tikzhl{punishred!80}{punish} the ``wrong'' attention scores towards the label tokens different from the query's ground-truth with magnitude $A\geqslant0$, and \tikzhl{rewardblue!80}{reward} the ``correct'' attention scores with magnitude $B\geqslant0$. These two terms in the loss function may seem redundant, but we will demonstrate in \S\ref{sec:loss_factor} that they actually contain antagonistic implicit biases, therefore, the factors $A$ and $B$ should be balanced well.


\paragraph{Why not End-to-end LoRA?} Intuitively, directly adding LoRA bypasses~\cite{hu2022lora} to the trained projection matrices and fine-tuning them on an end-to-end training objective is also a possible approach. However, in end-to-end LoRA, gradients are propagated from the output logits of LMs, which requires that the final layer of the LMs (i.e., output embeddings, or LM Head) must be in full precision and with gradients activated, to get stable gradients into the residual stream. This introduces a non-negligible overhead, an issue avoided by ABFT as it does not supervise the final output. Moreover, fine-tuning attention projections without selectivity may cause harmful side effects toward the ICL out of the fine-tuned domain. We will compare the performance and efficiency of ABFT against end-to-end LoRA in the following experiments (Table~\ref{table:e2eft}) to highlight the efficiency and harmlessness of the ABFT method.

\section{Main Experiments}

In this section, we mainly confirm the effectiveness of the proposed ABFT, and find that: ABFT effectively improves the ICL performance to about 10\%$\sim$20\% relatively, which requires the minimum parameters less than 0.05\% to be full precision and gradient, with other parameters free to be quantized and gradient-free, and utilize 0.01\% data cost compared to the previous works.

\subsection{Experiment Settings}
\label{sec:experiment_settings}

\paragraph{Models and Datasets.} We conduct our experiment on \textbf{9 modern LLMs}: GPT2 (Large, XL)~\cite{radford2019language}, Falcon3 7B~\cite{Falcon3}, Llama3 (8B, 43B, 56B)~\cite{llama3modelcard}, DeepSeek-R1 Distill Qwen 14B~\cite{deepseekai2025deepseekr1incentivizingreasoningcapability}, Qwen2.5 32B~\cite{qwen2.5, qwen2}, and SimpleScaling s1.1 32B~\cite{muennighoff2025s1simpletesttimescaling}; and \textbf{8 datasets}: SST2, SST5~\cite{SST2andSST5}, MR~\cite{MR}, \textbf{F}inancial \textbf{P}hrasebank~\cite{FP}, TREC~\cite{TREC1, trec2}, \textbf{Subj}ective~\cite{subjective}, \textbf{T}weet \textbf{E}val \textbf{E}motion~\cite{TEE}, \textbf{T}weet \textbf{E}val \textbf{H}ate~\cite{TEH} (Refer Appendix~\ref{appendix:model_datasets} for details). 

\paragraph{Hyperparameters.} We set: training samples $n_{d}=512$, the number of demonstrations per ICL sample $k=4$. A standard Adam optimizer~\cite{kingma2014adam} is used with learning rate $\mathrm{lr}=2\times10^{-5}$ and pseudo-batch-size $n_b=32$ (i.e., we average gradients per $n_b=32$ samples before performing a single gradient step). We set the initial values $A_0=0.5$, $B_0=1.0$, and dynamically balance them with the PID algorithm (refer to Appendix~\ref{appendix:PID}), stabilizing the number of attention heads identified as induction heads (see \S\ref{sec:loss_factor}). The models are trained for $n_{\text{step}}=32$ steps.

\paragraph{Quantization Settings.} Models over 10B are quantized to \texttt{4-bit}, with full-precision LoRA~\cite{hu2022lora} (inner dimension $r=16$) trained on $W_Q$ and $W_K$ with learning rate $10^{-4}$.

\paragraph{Baselines.} We compare with: Contextual Calibration with 512 training samples~\cite{zhao2021calibrate}, MetaICL end-to-end fine-tuning with 3.55M samples~\cite{min2022metaicl}, and PICL re-pre-training with 80M samples~\cite{gu-etal-2023-pre}.

\paragraph{Others.} We conduct all the experiments on a single NVIDIA A40 with 48GB VRAM. We repeat each experiment 4 times ($\leqslant$10B) or 2 times ($>$10B), and report the averaged results on 1024 fixed test inputs for each dataset. ICL-style inputs are built with library \textsc{StaICC}~\cite{cho2025staicc}. 

\subsection{Main Results}
\label{sec:main_res}

\input{Table/Tab2_withe2eft_res_old2}


\input{Table/robustness}

The test results are shown in Table~\ref{table:mainres}, where ABFT consistently outperforms all the baselines, even with enormous training sets (to MetaICL, $3.55\mathrm{M}/512\approx7000\times$) and full-model fine-tuning (remind that ABFT only focuses on the $W_Q$ and $W_K$ matrices), suggesting that ABFT is satisfyingly efficient in both time and data cost. Such results also provide strong empirical evidence for the effectiveness of induction heads in LLMs. 

\paragraph{Towards Mechanistic Controllability.} To the best of our knowledge, ABFT is the first approach to train models without accessing final outputs, enabling model controlling via intermediate features or activations. Through this practice, we prototypically implement one of the visions of mechanistic interpretability~\cite{rai2024practical}: by attributing the model’s inference to specific modules (\textit{circuits}), we enable their local optimization, thereby improving overall performance effectively and efficiently.

\begin{figure}
    \begin{minipage}[t]{0.482\linewidth}
        \flushright
        \includegraphics[width=1\linewidth]{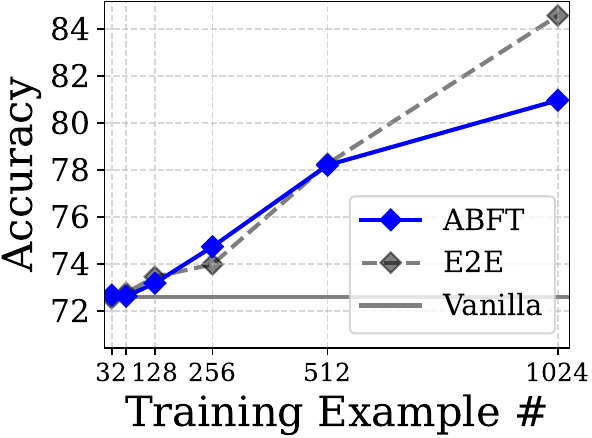}
        \vspace{-1.5\baselineskip}
        \caption{Data efficiency: accuracy against data size on DeepSeek-R1 Distill Qwen 14B / 8 datasets.}
        \label{fig:data_eff}
    \end{minipage} 
    \hspace{1.5pt}
    \begin{minipage}[t]{0.49\linewidth}
        \flushright
        \includegraphics[width=1\linewidth]{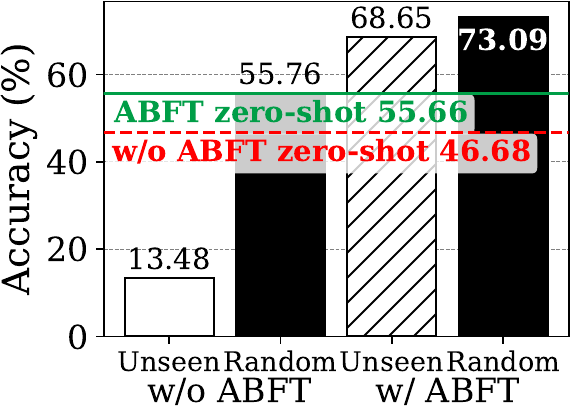}
        \vspace{-1.5\baselineskip}
        \caption{Accuracy on unseen label inputs and randomly sampled inputs, w/ and w/o ABFT.}
        \label{fig:unseen}
    \end{minipage}
    \vspace{-1\baselineskip}
\end{figure}

\paragraph{Prediction Consistency.} We evaluate the prediction consistency against variations in (1) prompt templates and (2) demonstration sampling on \textsc{StaICC-Diag}~\cite{cho2025staicc}. For each query, we repeat predictions across different prompt templates and sampling strategies, and measure consistency as the ratio of the maximum consistent predictions (e.g., 6 positive vs.\ 3 negative yields $6/9=2/3$), averaged over the dataset; see~\citet{cho2025staicc} and Appendix~\ref{appendix:stability} for implementation details. As shown in Table~\ref{table:robustness}, ABFT significantly improves consistency across all 8 datasets, stabilizing ICL under diverse contexts and enhancing prompt design efficiency. 

\paragraph{Prediction Bias against Wrong Labels.} Moreover, a known concern in ICL is the bias toward seen labels when ground-truth labels are absent in demonstrations (i.e., $\mathcal{I}^+=\emptyset$)~\cite{zhao2021calibrate, cho2025revisiting}, which can lead to incorrect predictions. Our testing on such scenario with and without ABFT in Fig.~\ref{fig:unseen} shows that (see Appendix~\ref{appendix:unseen_exp_method} for experiment details): ABFT mitigates this issue via the \tikzhl{punishred!80}{punish term $A$}, which penalizes incorrect labels during training and reduces the bias effects of induction heads. Notably, ABFT outperforms the 0-shot setting under unseen-label, suggesting the existence or emergence of unknown mechanisms that enable demonstrations in other categories to enhance ICL\footnote{Since that: such a phenomenon contrasts with existing views~\cite{cho2025revisiting}, where the explicit copying by the induction head is the only channel through which information is transferred from the demonstrations to the query (in Fig.~\ref{fig:unseen_attn_vis}, we show that the induction heads in ABFT model are almost fully suppressed in unseen-label scenario), where unseen-label demonstrations are harmful to ICL.}.

\subsection{Comparison against End-to-end Fine-tuning}

As mentioned before, end-to-end (E2E) fine-tuning on the in-domain dataset (not a wide dataset like MetaICL) with LoRA is also an alternative solution. However, in this section, we will show that compared with E2E fine-tuning, ABFT is more efficient, and more harmless on tasks out of the fine-tuning domain.

\paragraph{Time and Memory Cost.} Notice that in the E2E scenario, since the gradient is propagated from the final output of the model, the LM head, which is the top of the model, should be in full precision, to ensure a sufficient numerical resolution to utilize the mini-batch for mitigating the gradient noise in the stochastic gradient descent~\cite{hubara2016binarized}. This introduces a non-negligible overhead, as measured in Table~\ref{table:e2eft}. E2E fine-tuning slightly outperforms ABFT in in-domain accuracy (ACC$_\text{ID}$), but incurs substantial training time and memory costs.


\paragraph{Harmlessness.} We evaluate \textbf{o}ut-of-\textbf{d}omain (OD) performance on datasets different from the fine-tuned one (Table~\ref{table:e2eft}, ACC$_\text{OD}$). Both ABFT and E2E fine-tuning degrade OD performance, but ABFT causes less harm. This supports a conclusion from~\citet{cho2025revisiting}: some induction heads are intrinsic and task-independent, while others are task-induced. ABFT on intrinsic heads harms OD performance, whereas ABFT on task-induced heads does not. In contrast, E2E fine-tuning on all heads broadly degrades OD performance.

\paragraph{Data Efficiency.} We test the accuracy against the training set size as a metric of data efficiency, for both ABFT and E2E fine-tuning, as shown in Fig.~\ref{fig:data_eff} (refer to Appendix~\ref{appendix:more_res_on_DE} for results on other models). In the results, ABFT and E2E fine-tuning consistently benefit from more data samples, and in few-shot scenarios ($\leqslant512$), ABFT and E2E fine-tuning act equally, while E2E fine-tuning acts better when more training data is given. However, given the low-resource objective of ICL, and also the far more expensive time and memory cost of E2E fine-tuning, we can claim that ABFT has an advantage in the few-shot and low-resource scenario.

\begin{figure}[t]
    \centering
    \includegraphics[width=0.95\linewidth]{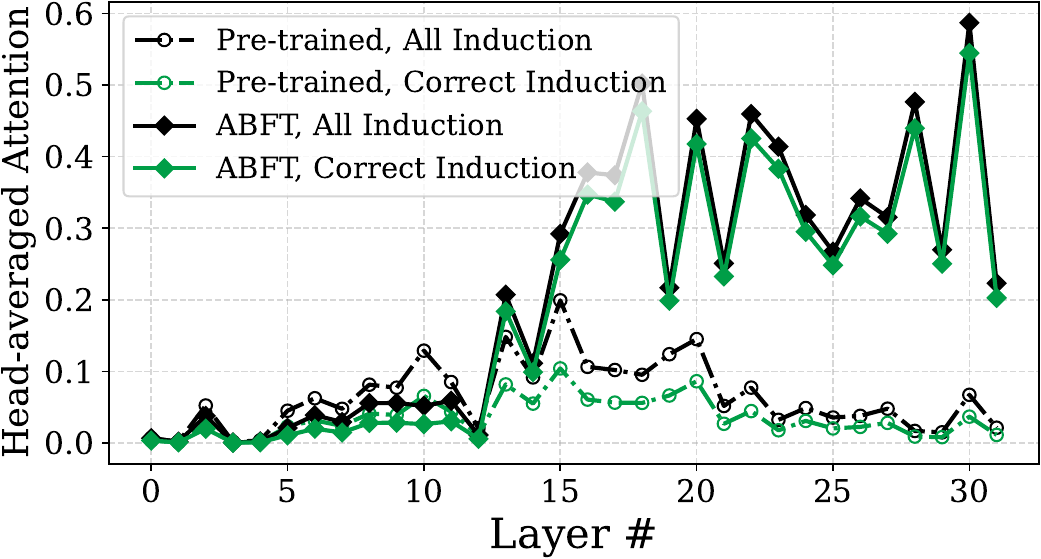}
    \vspace{-0.7\baselineskip}
    \caption{Induction attention scores and correct (on the correct labels) induction attention score averaged among heads on each layer. \textbf{ABFT enhances induction attention, and improves the correctness}.}
    \label{fig:attention_score}
    \vspace{-0.5\baselineskip}
\end{figure}

\section{Analysis}

\subsection{Attention Visualization after ABFT}

As shown in Fig.~\ref{fig:attention_score}, we average the global (to $\mathcal{I}$) and correct (to $\mathcal{I}^+$) induction attention scores on the last token among attention heads and input samples on each transformer layer, on the validation set. Also, we provide a direct visualization of attention scores in Appendix~\ref{appendix:attn_vis_more}. Compared to the pre-trained model, the ABFT model tends to eliminate attention scores towards incorrect label tokens ($\mathcal{I}^-$), and shift the attention scores from the attention sinks (the first token)~\cite{xiao2024efficient} and plain tokens to the correct label token, causing an enhancement to induction attention scores, continuously on the middle-to-last layers. Such visualization indicates that ABFT successfully generalizes to correct the behavior of attention heads.


\subsection{Ablation Analysis}
\label{sec:ablations}

In this section, we disable some components utilized in the ABFT training protocol to suggest their necessity. The main results of such ablation experiments are shown in Table~\ref{table:ablation}, where:

\input{Table/Tab3_Ablation}

\paragraph{ABFT should be Localized.} In Table~\ref{table:ablation}, disabling the head filter ((D) in Fig.~\ref{fig:main_figure}) harms the accuracy. Knowing that all attention heads are trained to be induction heads under unfiltered ABFT loss, we can infer that: in LLMs, some attention heads with functions other than the induction head are still necessary for ICL, aligning with and enhancing the previous work~\cite{cho2025revisiting}. However, considering that some implicit antagonistic effects induced by unfiltered ABFT loss still promote the formation of other essential heads (i.e., when the ABFT loss from deeper heads propagates to shallower heads, its function becomes antagonistic with the ABFT loss directly connected to those shallow heads), the accuracy degradation with no head filters is not so significant.

\paragraph{Loss Factor ($A$ and $B$) should be Balanced.} As mentioned in~\S\ref{sec:experiment_settings}, we use the PID algorithm to adaptively balance the value of $A$ and $B$ in the loss calculation. In Table~\ref{table:ablation}, we disable such adjustment and observe an accuracy drop. Especially, when we set the $A$ or $B$ to $0$, the accuracy significantly degrades. This eliminates the doubt of ``whether the loss factors are redundant'' raised in \S\ref{sec:method}, and we will discuss this in-depth in the next section (\S\ref{sec:loss_factor}).

\begin{figure}[t]
    \centering
    \includegraphics[width=0.95\linewidth]{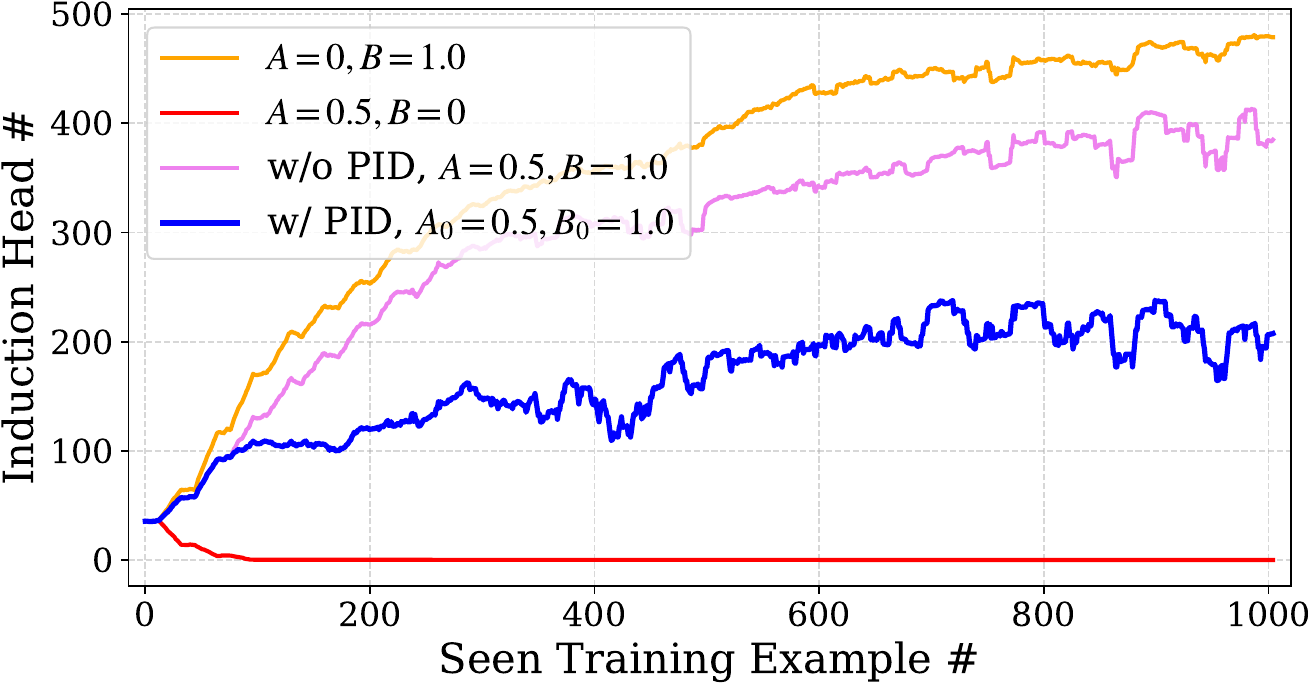}
    \vspace{-0.5\baselineskip}
    \caption{Number of attention heads that are judged as induction heads on 4 settings, against the training processing (the number of seen data samples). \textbf{PID effectively stabilizes the induction head numbers}.}
    \label{fig:induction_head_number}
    \vspace{-0.5\baselineskip}
\end{figure}

\subsection{Balance the Loss Factor $A$ and $B$}
\label{sec:loss_factor}


\begin{figure}
    \centering
    \includegraphics[width=\linewidth]{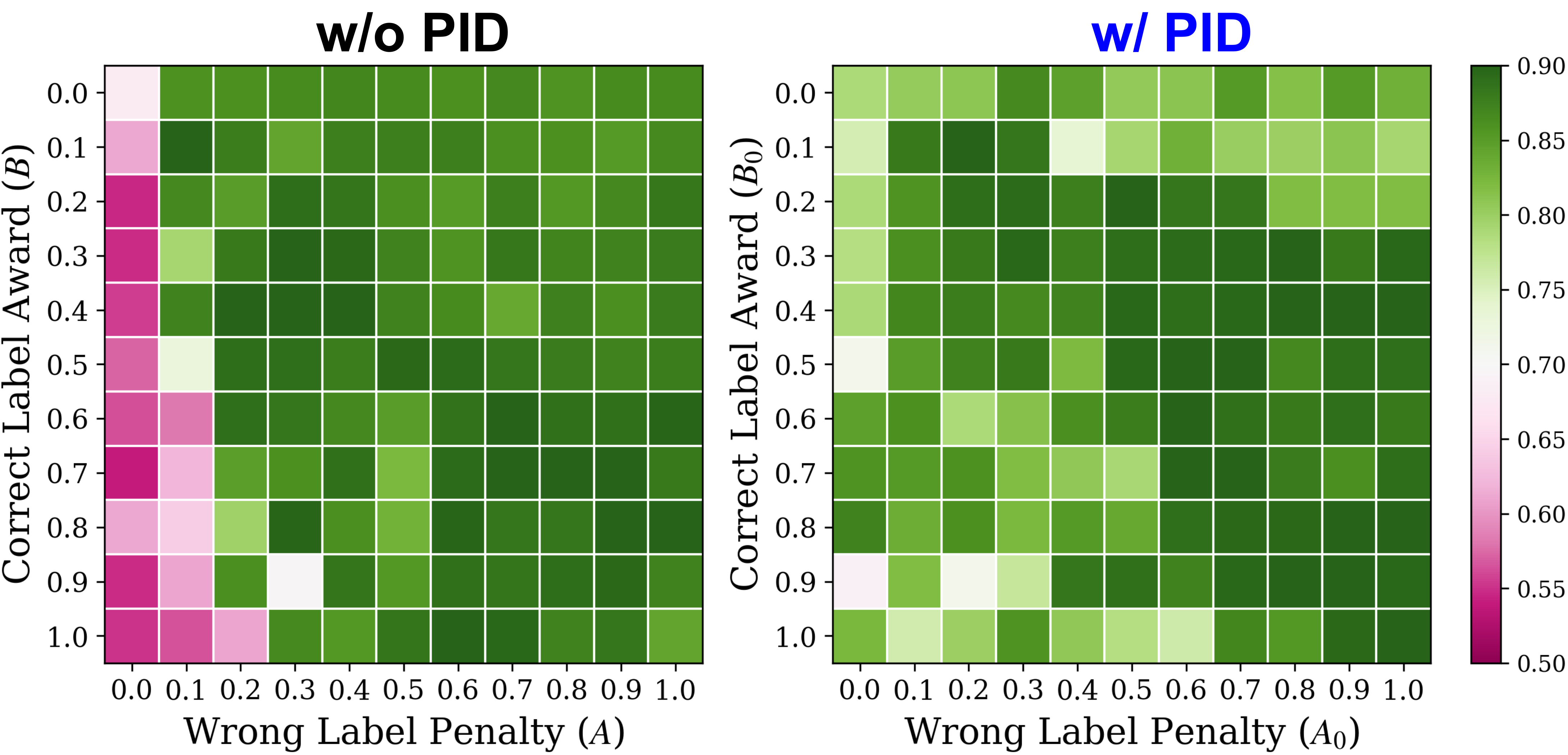}
    \vspace{-1.5\baselineskip}
    \caption{Accuracies with various settings on hyperparameter $A$ and $B$, with and without PID algorithm. \textbf{PID weakens the sensitivity to initial parameters}.}
    \label{fig:hyperparameter_map}
    \vspace{-0.5\baselineskip}
\end{figure}


\paragraph{\tikzhl{punishred!80}{Punish} and \tikzhl{rewardblue!80}{Reward} Influence Induction Heads Antagonisticly.} As shown in Table~\ref{table:ablation}, removing either loss term in Eq.~\ref{eq:ABFTLoss} degrades performance, indicating that both are essential. Interestingly, the two terms introduce antagonistic implicit biases: the \tikzhl{punishred!80}{punish term $A$} disperses attention across labels, reducing induction heads, while the \tikzhl{rewardblue!80}{reward term $B$} concentrates attention on specific labels, increasing induction heads through a stepwise positive feedback loop. We track the number of induction heads during training on Llama3 8B and SST2 (see Appendix~\ref{appendix:attn_head_n_more} for more cases), as shown in Fig.~\ref{fig:induction_head_number}, to support this observation.

\begin{figure}
    \centering
    \includegraphics[width=0.55\linewidth]{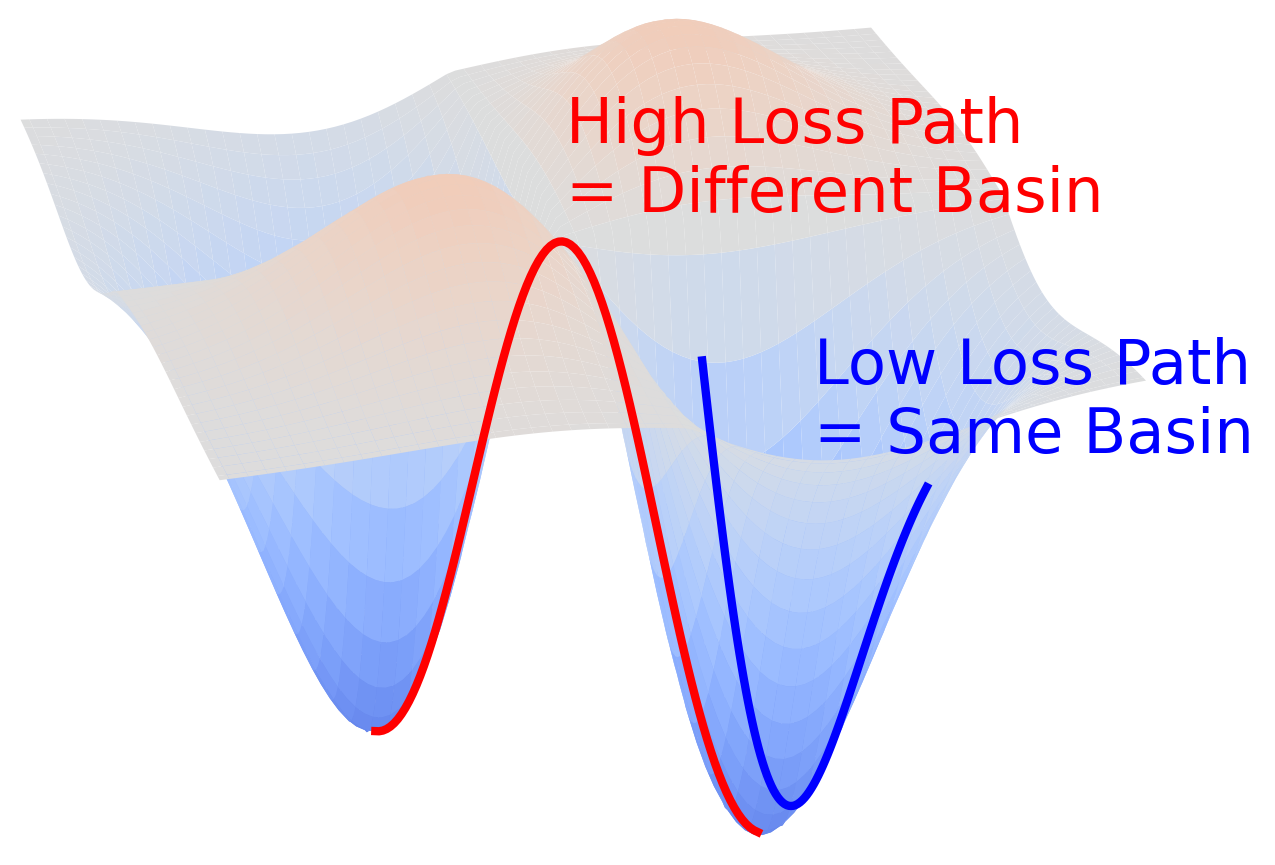}
    \includegraphics[width=0.9\linewidth]{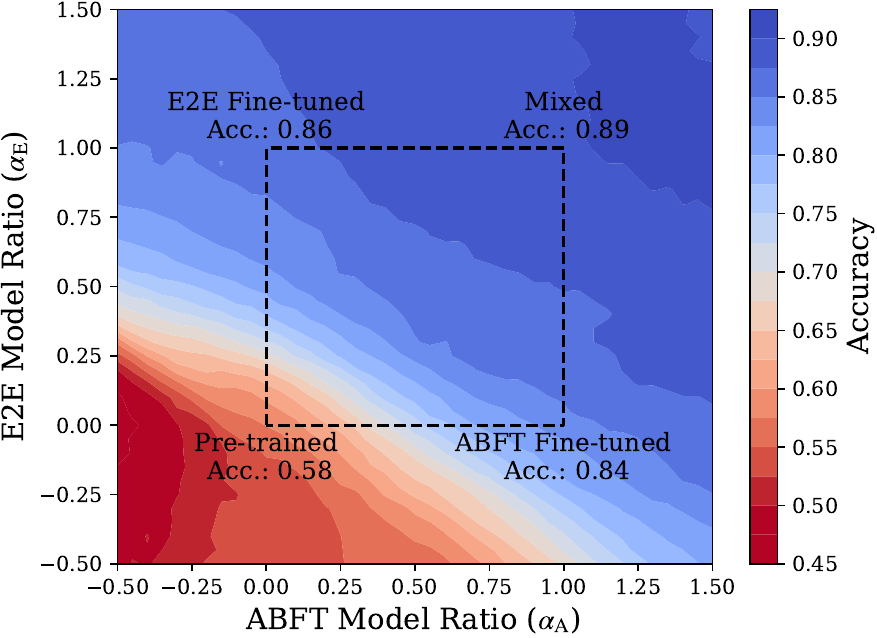}
    \vspace{-0.5\baselineskip}
    \caption{\textbf{Upper}: A better accuracy (lower loss) in the interpolation path suggests the same basin, and vice versa. \textbf{Lower}: Contour map of accuracies against the coefficient $\alpha_\text{E}$ and $\alpha_\text{A}$ in Eq.~\ref{eq:2}. \textbf{E2E fine-tuned and ABFT models are located in the same low-loss area.}}
    \label{fig:mix}
    \vspace{-0.5\baselineskip}
\end{figure}


\paragraph{Automatically Stabilizing Induction Head Number.} Ablation studies reveal that too many induction heads hinder fine-tuning and overall performance, as other functional heads are also needed for ICL; whereas an insufficient number prevents the model from handling ICL tasks. To maintain a stable number, we automatically adjust the antagonistic factor~$A$ in Eq.~\ref{eq:ABFTLoss} using a classical PID controller (Appendix~\ref{appendix:PID})\footnote{Since $A$ and $B$ are antagonistic, controlling $A$ alone suffices to stabilize induction head numbers.}. As shown in Fig.~\ref{fig:induction_head_number}, PID stabilizes induction head count, improves ABFT performance (Table~\ref{table:ablation}), and reduces sensitivity to hyperparameters $A_0$ and $B_0$, with accuracy remaining stable across settings (Fig.~\ref{fig:hyperparameter_map}).

\section{Consistency of Training Objective:\\ \hspace{1.25em} ABFT and End-to-end Fine-tuning}

To explore whether the emergence of induction head is from ICL-style data—a key question in interpretability~\cite{chan2022data, reddy2023mechanistic, singh2024needs}—we examine the consistency between ABFT and E2E objectives.

\begin{figure}
    \centering
    \includegraphics[width=1\linewidth]{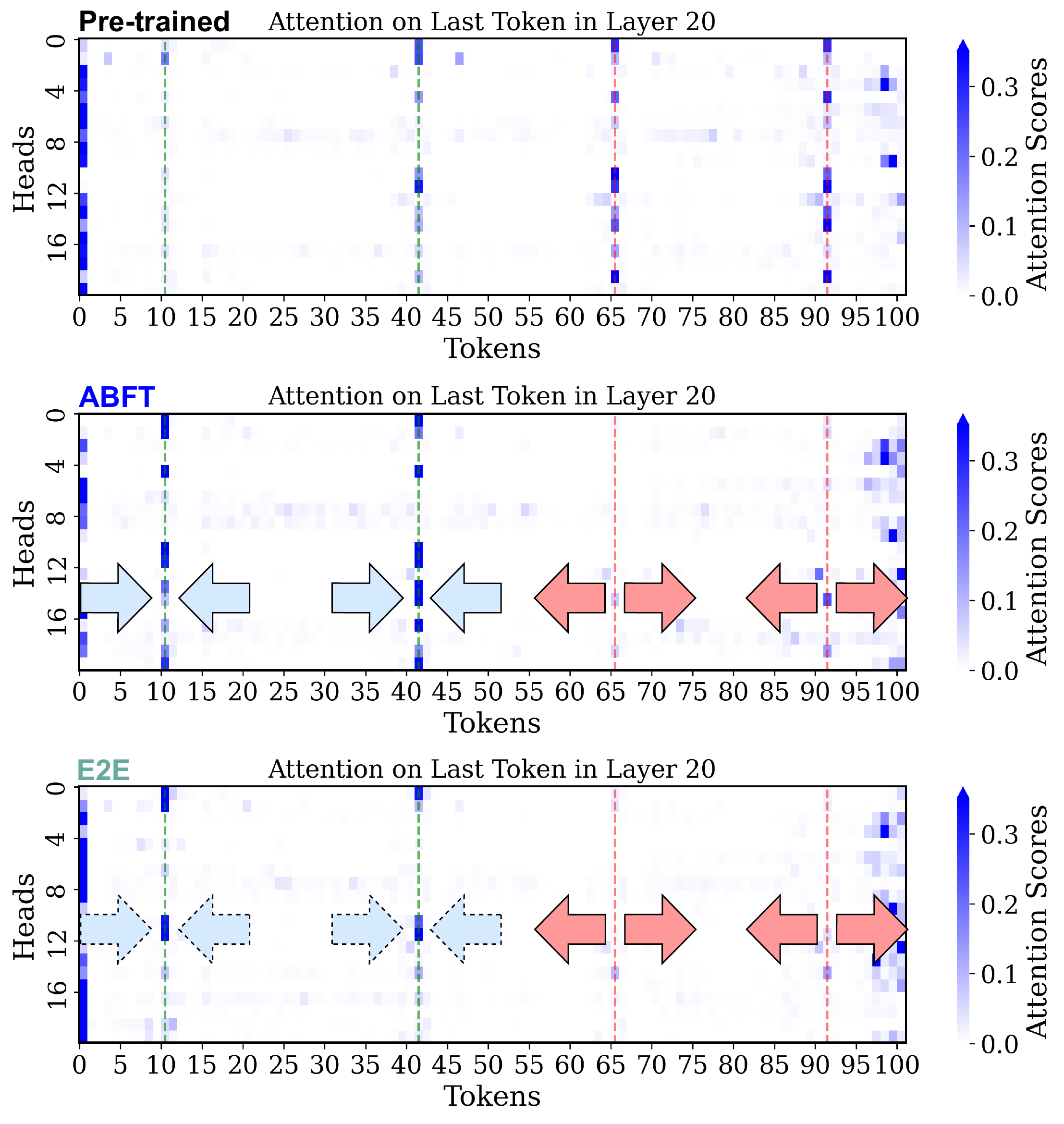}
    \vspace{-1.5\baselineskip}
    \caption{Attention score visualization of the pre-trained model, ABFT model, and E2E fine-tuned model, on the same input as Fig.~\ref{fig:attention_visual}, and the same models as Fig.~\ref{fig:mix}  (Refer to Appendix~\ref{appendix:attn_vis_more} for details and more cases).}
    \label{fig:attention_visualization_3}
    \vspace{-0.5\baselineskip}
\end{figure}

\paragraph{Principle.} Due to the lack of qualitative thresholds, it is hard to utilize statistic-based similarity measures to determine whether two models exhibit comparable similarity sufficient to indicate consistent training objectives. Therefore, our experiment is based on such a principle: if the fine-tuning terminations on both training objectives fall into the same \textit{basin} of the loss function, then both fine-tuning trajectories are similar~\cite{neyshabur2020being}, so that the two training objectives are consistent. For such an end, we investigate the \textit{linear-connectivity}~\cite{neyshabur2020being, ilharcoediting} among the pre-trained parameters $\theta_{0}$ (as the start point of the fine-tuning) and fine-tuned parameters $\theta_{\text{E}}$ for E2E fine-tuning, and $\theta_{\text{A}}$ for ABFT. In detail, we mix these three parameters into a new model parameter set $\theta$ in the following form:
\begin{equation}
    \theta = \theta_0+\alpha_\text{E}(\theta_\text{E}-\theta_0)+\alpha_\text{A}(\theta_\text{A}-\theta_0),
    \label{eq:2}
\end{equation}
\noindent and then test the accuracy of $\theta$ as an anti-metric of model loss. As shown in Fig.~\ref{fig:mix} (upper), if the accuracy of mixed $\theta$ is better (or at least, not significantly worse) than the accuracy of $\theta_{\text{E}}$ and $\theta_{\text{A}}$, we can infer that the $\theta_{\text{E}}$ and $\theta_{\text{A}}$ are in the same loss basin, with linear low-loss path observed.

\begin{figure}
    \centering
    \includegraphics[width=0.9\linewidth]{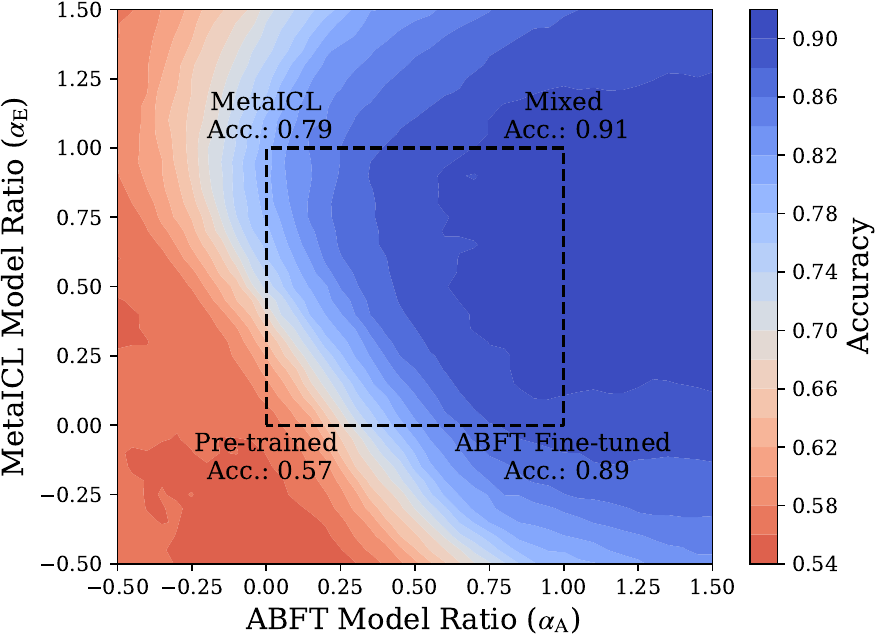}
    \vspace{-0.5\baselineskip}
    \caption{Contour map of accuracies against the $\alpha_\text{E}$ and $\alpha_\text{A}$ on GPT2-L, and the $\theta_\text{E}$ is set as MetaICL model.}
    \label{fig:mix_meta}
    \vspace{-0.5\baselineskip}
\end{figure}

\paragraph{Experiment and Result.} We conduct the aforementioned experiment protocol on SST2 and GPT2-XL. The results are shown in Fig.~\ref{fig:mix} (lower), showing no high-loss paths between the E2E fine-tuned model and the ABFT model. It suggests that they are in the same basin of the loss landscape, indicating the high similarity between these two training objectives. Moreover, we visualize the attention scores on ABFT and E2E fine-tuned models as shown in Fig.~\ref{fig:attention_visualization_3}: compared with the pre-trained model, attention scores of the fine-tuned models on both objectives consistently focus on the correct label tokens, suggesting that E2E objectives imply a promotion to correct induction head. Moreover, as shown in Fig.~\ref{fig:mix_meta}, repeating the experiment on MetaICL shows MetaICL model lies in the same basin as ABFT, suggesting that full-model tuning on large datasets is essentially equivalent to ABFT, which can be seen as \textit{essential ICL fine-tuning}.

\section{Discussion}

\paragraph{Conclusion.} In this paper, we propose a fine-tuning objective that strengthens the correctness of the induction head by accessing only the attention matrix, and demonstrate that it significantly improves the performance of ICL. Our results reinforce the induction head hypothesis for ICL interpretability and represent a first step toward controlling model behavior through mechanistic interpretability.

\paragraph{Towards Mechanistic Controllability.} In this paper, we raise the possibility of controlling the model's behavior by some specific modules (often called \textit{circuit} in the context of mechanistic interpretability), which opens up a new neural network model behavior-controlling paradigm: controlling only the modules that make significant contributions to the output, thereby substantially reducing the number of parameters that need to be adjusted and achieving excellent efficiency.

\section*{Limitations} 
\paragraph{Towards Open-end Tasks.} The first limitation lies in the fact that the induction head-based explanation of ICL~\cite{elhage2021mathematical, singh2024needs, cho2025revisiting}, so that our proposed ABFT approach, applies only to classification tasks with a finite label set. As mentioned in \S\ref{sec:main_res} and Fig.~\ref{fig:unseen}, since our training objective consists of two factors, our approach is not limited to the simple retrieval setting where the ground-truth label appears in the demonstrations, while extending these methods to open-ended tasks remains an open challenge that requires further investigation on the basic mechanism. Nevertheless, given the current state of research on ICL interpretability, we have made full use of these findings and provided a valuable foundation for advancing model control through the scope of interpretability, i.e., Mechanistic Controllability.

\paragraph{Towards Better Mechanistic Controllability.} For our vision of Mechanistic Controllability, even though this paper successfully identifies a small set of modules (i.e., circuits) that require controlling towards better ICL performance, the control methods based on gradients and moderate amounts of data remain coarse. Therefore, future work could focus on gradient-free and data-free model editing, which directly edits some parameters utilizing a deeper understanding of the functional roles of model parameters.

\paragraph{Towards Better Performance.} It can be considered that some hyperparameters (see \S\ref{sec:experiment_settings} and Appendix~\ref{appendix:PID}), and the induction head filter (see \S\ref{sec:method}) may be not optimal, restricting the performance. Discussing them in detail, and automatically optimizing them can be helpful for better performance of ABFT. Also, in Fig.~\ref{fig:mix}, we observe that the extended line from the pre-trained model towards the ABFT model leads to better accuracy, suggesting a possibility of utilize model parameter $\theta = \theta_0+\alpha_\text{A}(\theta_\text{A}-\theta_0), \alpha_\text{A}>1$ to further improve accuracy without any gradient-based cost.

\paragraph{Towards Further Efficiency.} As shown in Appendix~\ref{appendix:parameter_shift}, the $W_Q$ and $W_K$ projections are significantly modified after ABFT only in some layers, that is, it is possible to further restrict the gradient-on parameters to some Transformer layers for better efficiency (notice that currently we activate the gradients of the attention mappings of all layers).

\newpage


\subsection*{Acknowledgements}


This work was supported by JST FOREST Program (Grant Number JPMJFR232K, Japan) and the Nakajima Foundation. 

\newpage

\bibliography{custom}

\input{appendix}

\end{document}

%% file: Table/Tab1_mainres.tex
\begin{table*}[t]
\centering
\caption{Accuracies (\%) of ABFT and baselines on 9 LMs and 8 datasets. The best results are in \textbf{bold}.}
\vspace{-0.5\baselineskip}
\label{table:mainres}
\resizebox{\textwidth}{!}{
\begin{tabular}{@{}cclcccccccc|c@{}}
\toprule
\textbf{Model} &
  \textbf{Param.} &
  \textbf{Method} &
  {SST2} &
  {MR} &
  {FP} &
  {SST5} &
  {TREC} &
  {SUBJ} &
  {TEE} &
  {TEH} &
  \textbf{Average} \\ \midrule
\multirow{5}{*}{\textbf{GPT2-L}}  & \multirow{5}{*}{812M} & Vanilla        & 56.35 & 61.13 & 42.09 & 29.69 & 35.45 & 49.12 & 39.06 & 54.00 & 45.86 \\
                                  &                       & MetaICL        & 85.94 & 80.96 & 37.30 & \textbf{42.09} & 33.98 & 50.49 & 45.41 & 54.20 & 53.80 \\
                                  &                       & PICL           & 74.70 & 73.34 & 54.49 & 33.79 & 32.91 & 51.37 & \textbf{47.46} & 53.42 & 52.68 \\
                                  &                       & Calibrate      & 56.35 & 60.94 & 36.91 & 25.10 & 34.08 & 49.32 & 36.72 & 53.52 & 44.12 \\ 
                                  &                       & \textbf{ABFT}  & \textbf{88.18}\std{1.54} & \textbf{85.40}\std{0.95} & \textbf{81.30}\std{1.69} & 36.84\std{3.61} & \textbf{50.24}\std{2.49} & \textbf{61.99}\std{4.39} & 46.51\std{3.11} & \textbf{55.20}\std{0.20} & \textbf{63.21} \\ \midrule

\multirow{4}{*}{\textbf{GPT2-XL}} & \multirow{4}{*}{1.61B} & Vanilla       & 67.87 & 69.53 & 51.07 & 30.66 & 35.25 & 50.98 & 42.68 & 53.03 & 50.13 \\
                                 &                         & PICL          & 74.80 & 74.32 & 51.17 & 32.71 & 33.20 & 51.46 & 47.95 & 53.42 & 52.38 \\
                                  &                       & Calibrate      & 68.16 & 75.00 & 36.43 & 28.52 & 35.55 & 50.10 & 39.26 & 51.56 & 48.07 \\
                                  &                       & \textbf{ABFT}  & \textbf{87.92}\std{1.47} & \textbf{86.52}\std{1.50} & \textbf{87.67}\std{0.45} & \textbf{37.55}\std{2.67} & \textbf{51.83}\std{2.73} & \textbf{75.07}\std{2.96} & \textbf{60.01}\std{5.38} & \textbf{55.35}\std{0.04} & \textbf{67.74} \\ \midrule
                                  
\multirow{3}{*}{\textbf{Falcon3}} & \multirow{3}{*}{7.46B} & Vanilla       & 91.11 & 92.77 & 85.35 & 46.00 & 50.00 & 62.60 & 60.55 & 52.05 & 67.55 \\
                                  &                       & Calibrate      & 90.53 & \textbf{93.07} & 82.71 & 44.04 & 54.30 & 62.40 & 54.79 & 51.76 & 66.70 \\
                                  &                       & \textbf{ABFT}  & \textbf{92.14}\std{0.21} & 92.17\std{0.04} & \textbf{96.14}\std{0.36} & \textbf{47.32}\std{0.16} & \textbf{75.81}\std{0.19} & \textbf{94.87}\std{0.82} & \textbf{67.97}\std{0.25} & \textbf{70.34}\std{0.22} & \textbf{79.59} \\ \midrule
                                  
\multirow{3}{*}{\textbf{Llama3}}  & \multirow{3}{*}{8.03B} & Vanilla        & 89.35 & 92.87 & 75.78 & 44.24 & 55.76 & 62.30 & 57.91 & 54.59 & 66.60 \\
                                  &                       & Calibrate      & 90.04 & \textbf{93.36} & 43.95 & 41.60 & 54.39 & 65.23 & 54.79 & 52.83 & 62.02 \\
                                  &                       & \textbf{ABFT}  & \textbf{93.14}\std{1.39} & 92.50\std{0.84} & \textbf{94.02}\std{1.72} & \textbf{52.10}\std{2.31} & \textbf{73.09}\std{1.11} & \textbf{92.70}\std{1.44} & \textbf{72.02}\std{1.81} & \textbf{72.02}\std{5.87} & \textbf{80.20} \\ \midrule
                                  
\multirow{3}{*}{\thead{\textbf{DeepSeek-R1} \\Dist.\ Qwen\\ \texttt{4-bit}, LoRA}}    & \multirow{3}{*}{14.8B} 
                                                          & Vanilla        & 90.92 & 91.21 & 92.18 & \textbf{46.97} & 62.50 & 66.60 & 66.60 & 63.87 & 72.61 \\
                                  &                       & Calibrate      & 90.04 & 91.41 & \textbf{92.68} & 46.09 & 61.62 & 65.43 & 65.33 & 62.30 & 71.86 \\
                                  &                       & \textbf{ABFT}  & \textbf{93.51}\std{1.22} & \textbf{91.85}\std{0.34} & 91.17\std{4.44} & 46.09\std{1.27} & \textbf{69.14}\std{2.54} & \textbf{92.92}\std{3.27} & \textbf{69.82}\std{0.00} & \textbf{71.19}\std{3.42} & \textbf{78.21} \\ \midrule
                                  
\multirow{3}{*}{\thead{\textbf{Qwen2.5}\\ \texttt{4-bit}, LoRA}}    & \multirow{3}{*}{32.8B} 
                                                          & Vanilla        & 93.85 & 94.43 & 86.23 & 47.17 & 58.40 & 87.50 & 65.14 & 69.63 & 75.29 \\
                                  &                       & Calibrate      & 93.75 & 94.82 & 74.22 & 44.82 & 58.79 & 84.08 & 63.96 & 63.96 & 72.30 \\
                                  &                       & \textbf{ABFT}  & \textbf{94.92}\std{0.00} & \textbf{94.83}\std{0.10} & \textbf{94.04}\std{0.00} & \textbf{48.49}\std{0.05} & \textbf{69.24}\std{0.10} & \textbf{96.00}\std{0.00} & \textbf{69.29}\std{0.24} & \textbf{70.31}\std{0.00} & \textbf{79.64} \\ \midrule
                                  
\multirow{3}{*}{\thead{\textbf{SimpleScaling}\\ s1.1 \\ \texttt{4-bit}, LoRA}}    & \multirow{3}{*}{32.8B} 
                                                          & Vanilla        & 94.82 & 94.24 & 91.11 & \textbf{50.20} & 69.63 & 89.65 & 68.36 & 72.17 & 78.77 \\
                                  &                       & Calibrate      & 94.43 & 93.85 & 88.96 & 48.63 & 68.07 & 89.26 & 68.55 & 72.36 & 78.02 \\
                                  &                       & \textbf{ABFT}  & \textbf{94.92}\std{0.10} & \textbf{94.29}\std{0.05} & \textbf{96.00}\std{0.00} & 49.95\std{0.05} & \textbf{72.46}\std{0.39} & \textbf{95.71}\std{0.10} & \textbf{71.73}\std{0.05} & \textbf{73.10}\std{0.04} & \textbf{81.02} \\ \midrule
                                  
\multirow{3}{*}{\thead{\textbf{Llama3}\\ \texttt{4-bit}, LoRA}}    & \multirow{3}{*}{43.2B} 
                                                          & Vanilla        & 93.26 & 94.04 & 73.92 & \textbf{49.41} & 58.98 & 71.58 & 62.60 & 66.70 & 71.31 \\
                                  &                       & Calibrate      & \textbf{95.02} & 93.07 & 54.20 & 44.53 & 59.08 & 72.56 & 61.03 & 65.82 & 68.16 \\
                                  &                       & \textbf{ABFT}  & \textbf{95.02}\std{0.10} & \textbf{93.85}\std{0.10} & \textbf{94.87}\std{0.05} & 48.10\std{0.14} & \textbf{64.70}\std{0.14} & \textbf{90.09}\std{0.05} & \textbf{69.24}\std{0.29} & \textbf{70.85}\std{0.15} & \textbf{78.34} \\ \midrule
                                  
\multirow{3}{*}{\thead{\textbf{Llama3}\\ \texttt{4-bit}, LoRA}}    & \multirow{3}{*}{55.6B} 
                                                          & Vanilla        & 93.94 & 92.19 & 78.81 & \textbf{51.37} & 67.19 & 66.70 & 56.44 & 60.94 & 70.95 \\
                                  &                       & Calibrate      & 92.29 & 92.77 & 69.92 & 50.49 & 68.75 & 65.92 & 58.11 & 62.70 & 70.12 \\
                                  &                       & \textbf{ABFT}  & \textbf{94.53}\std{0.10} & \textbf{93.41}\std{0.14} & \textbf{93.21}\std{0.44} & 49.02\std{0.78} & \textbf{71.78}\std{0.58} & \textbf{92.68}\std{0.48} & \textbf{70.76}\std{0.64} & \textbf{70.75}\std{0.34} & \textbf{79.52} \\ \bottomrule
\end{tabular}}
\end{table*}

%% file: Table/Tab2_withe2eft_res_old2.tex
\begin{table}[t]
\centering
\caption{A comparison between ABFT and End-to-end Fine-tuning (E2E FT). Param.*: Parameters which are required in \texttt{FP16/32} and with gradient on.}
\vspace{-0.5\baselineskip}
\label{table:e2eft}
\resizebox{\linewidth}{!}{
\begin{tabular}{@{}clcccc@{}}
\toprule
\textbf{Model} &
  \textbf{Method} &
  \textbf{Param.*} &
  \textbf{Time} &
  \textbf{Acc${}_{\text{ID}}$} &
  \textbf{Acc${}_{\text{OD}}$} \\ \midrule

\multirow{3}{*}{\thead{\textbf{Llama3} \\8.03B \\ \texttt{4-bit}, LoRA}} 
                                                          & Vanilla & - & - & \multicolumn{2}{c}{66.02}\\
                                  & E2E FT  & 0.5B & $2.2\times$ & 78.33 & 61.74\\
                                  & \textbf{ABFT}  & 6.8M & $1\times$ & 72.54 & 64.34\\ \midrule
                                  
\multirow{3}{*}{\thead{\textbf{DeepSeek-R1} \\14.8B \\ \texttt{4-bit}, LoRA}} 
                                                          & Vanilla & - & - & \multicolumn{2}{c}{72.61}\\
                                  & E2E FT  & 0.8B & $2.2\times$ & 78.26 & 63.62 \\
                                  & \textbf{ABFT}  & 12M & $1\times$ & 78.21 & 67.21 \\ \midrule
\multirow{3}{*}{\thead{\textbf{Qwen2.5} \\32.8B \\ \texttt{4-bit}, LoRA}} 
                                                          & Vanilla & - & - & \multicolumn{2}{c}{75.29} \\
                                  & E2E FT  & 0.8B & $2.6\times$ & 82.09 & 62.24 \\
                                  & \textbf{ABFT}  & 17M & $1\times$ & 79.64 & 64.96 \\ \midrule
\multirow{3}{*}{\thead{\textbf{Llama3} \\ 55.6B \\ \texttt{4-bit}, LoRA}} 
                                                          & Vanilla & - & - & \multicolumn{2}{c}{70.95} \\
                                  & E2E FT  & 1.1B & $2.7\times$ & 82.80 & 64.86 \\
                                  & \textbf{ABFT}  & 33M & $1\times$ & 79.52 & 67.32 \\ \bottomrule
\end{tabular}}
\vspace{-0.4\baselineskip}
\end{table}


%% file: Table/robustness.tex
\begin{table}[t]
\caption{Prediction consistency metrics (\%) on each models averaged among 8 datasets.}
\label{table:robustness}
\vspace{-0.5\baselineskip}
\resizebox{\linewidth}{!}{
\begin{tabular}{@{}ccccc@{}}
\toprule
\multirow{2}{*}{\textbf{Model}} & \multicolumn{2}{c}{\textbf{Template Consist.}} & \multicolumn{2}{c}{\textbf{Demonstration Consist.}} \\ \cmidrule(l){2-3} \cmidrule(l){4-5}
                       & w/o ABFT       & w/ ABFT       & w/o ABFT       & w/ ABFT       \\ \midrule
GPT2-XL                &   81.28        &  \textbf{91.74}     &    68.38     &   \textbf{82.75}        \\
Llama3 8B              & 86.93          & \textbf{90.32}         & 76.99          & \textbf{92.00}         \\
DeepSeek-R1 14B        & 89.64          & \textbf{92.79}         & 81.30          & \textbf{85.97}         \\
Qwen2.5 32B            & 88.97          & \textbf{92.78}         & 84.52          & \textbf{87.94}         \\
Llama3 56B             & 92.78          & \textbf{93.90}         & 82.10          & \textbf{87.49}         \\ \bottomrule
\end{tabular}}
\vspace{-0.7\baselineskip}
\end{table}

%% file: Table/Tab3_Ablation.tex
\begin{table}[t]
\centering
\caption{Ablation analysis of removing some components from ABFT. Notice that the PID algorithm is to stabilize the induction head number by adjusting the factor $A$, so when we disable the head filter or fix the $A$ or $B$, the PID algorithm naturally loses its function.}
\vspace{-0.5\baselineskip}
\label{table:ablation}
\resizebox{0.95\linewidth}{!}{
\begin{tabular}{@{}clcc@{}}
\toprule
\textbf{Model} &
  \textbf{Method} &
  \textbf{Time} &
  \textbf{Acc.} \\ \midrule
\multirow{6}{*}{\thead{\textbf{Falcon3}\\ 7.46B}} & Vanilla & - & 67.55 \\
                                & \textbf{ABFT} & $1\times$ & \textbf{79.59} \\ 
                                  & w/o PID, $A=0.5, B=1.0$ & $1.0\times$ & 76.86 \\
                                   & w/o PID, w/o Head Filter & $1.3\times$ & 75.57 \\
                                   & w/o PID, $A=0, B=1.0$ & $1.0\times$ & 72.13 \\ 
                                   & w/o PID, $A=0.5, B=0$ & $0.6\times$ & 56.47 \\ \midrule
                                  
\multirow{6}{*}{\thead{\textbf{Llama3}\\ 8.03B}} & Vanilla & - & 66.60 \\
                                & \textbf{ABFT} & $1\times$ & \textbf{80.20} \\ 
                                  & w/o PID, $A=0.5, B=1.0$ & $1.1\times$ & 80.07 \\
                                   & w/o PID, w/o Head Filter & $1.2\times$ & 70.39 \\
                                   & w/o PID, $A=0, B=1.0$ & $1.2\times$ & 63.54 \\ 
                                   & w/o PID, $A=0.5, B=0$ & $0.6\times$ & 58.79 \\ \midrule
\multirow{6}{*}{\thead{\textbf{DeepSeek-R1} \\14.8B \\Dist.\ Qwen\\ \texttt{4-bit}, LoRA}}  & Vanilla & - & 72.61 \\
                                                         &  \textbf{ABFT}  & $1\times$ & \textbf{78.21} \\ 
                                  & w/o PID, $A=0.5, B=1.0$& $1.0\times$ & 73.35 \\
                                   & w/o PID, w/o Head Filter & $2.1\times$ & 73.51 \\
                                   & w/o PID, $A=0, B=1.0$ & $0.9\times$ & 72.71 \\ 
                                   & w/o PID, $A=0.5, B=0$ & $0.9\times$ & 73.36 \\ \bottomrule
\end{tabular}}
\vspace{-0.5\baselineskip}
\end{table}

%% file: appendix.tex
\newpage

\appendix

\section{Detailed Experiment Implementation}

\subsection{Model \& Dataset Details}
\label{appendix:model_datasets}

\input{Table/prompt_templates}

\input{Table/model_repo}

\paragraph{Models.} All the models in this paper are loaded from \texttt{huggingface}. In detail, we list the \texttt{huggingface} repository name to keep the repeatability of this paper, as shown in Table~\ref{appendix.tab:repo}.

\paragraph{Dataset Split.} As also described by~\citet{cho2025staicc}, we randomly sample 1024 data samples from the original dataset to build the inputs for training, and sample $4096+512$ (especially, $512+512$ for FP dataset, $3192+512$ for TEH dataset) data samples for the demonstrations$+$queries for the testing, respectively.

\paragraph{Demonstration Sampling.} To generate the training examples of $k$ demonstrations, we randomly sample $(k+1)$ data examples from the aforementioned 1024 data, and concatenate them into the inputs, with the prompt templates shown in \S\ref{appendix:prompt_temp}. To generate the testing examples, for each query in the 512 samples, we sample two sequences of demonstrations from the 4096 data samples, and concatenate them into testing inputs, 2 for one query sample, so that 1024 for one dataset.

\subsection{Prompt Templates}
\label{appendix:prompt_temp}

We utilize the default prompt templates of \textsc{StaICC}, as shown in Table~\ref{tab:prompt_templates}. For the sake of simplicity, we reduce the label tokens into one token, as also shown in Table~\ref{tab:prompt_templates}.

\subsection{Details of PID Algorithm}
\label{appendix:PID}

On each model update step $t>2$ (i.e., when the gradients from all the samples of the $t$-th pseudo batch (of $n_b$ data samples) are propagated), we calculated the identified induction head numbers from the induction head filter described in \S\ref{sec:method} and Fig.~\ref{fig:main_figure} averaged on the $n_b$ data samples as $\bar{n}_{t}$. Given the similar averaged induction head number on the previous time step $(t-1)$ as $\bar{n}_{t-1}$, we can calculate the updated $A_{t}$ term\footnote{Remind that for the sake of simplicity, we only control $A$, given the findings of $A$ and $B$ are antagonistic, as shown in \S\ref{sec:loss_factor}.} by standard PID algorithm as:
\begin{equation}
\begin{split}
    A_{t} &= C_p\left(\bar{n}_{t} - \bar{n}_{t-1}\right) \\ &+ C_i\left(\sum_{i=2}^t\bar{n}_{i} - \bar{n}_{i-1}\right) \\ &+
    C_d\left(\bar{n}_{t} - 2\bar{n}_{t-1} + \bar{n}_{t-2}\right) \\ &+
    A_{t-1},
\end{split}
\label{eq:PID}
\end{equation}
\noindent where the $C_p = 0.03$, $C_i=0.005$, and $C_d=0.005$ are hyperparameters. By such calculation, we implement a feedback control to stabilize the number of induction heads among training steps. 

\subsection{Experiment Protocol of Unseen Label}
\label{appendix:unseen_exp_method}

In Fig.~\ref{fig:unseen}, we examine that ABFT model can utilize the demonstration with wrong label to improve the ICL performance. Here we introduce the experiment protocol.

\begin{figure}
    \centering
    \includegraphics[width=\linewidth]{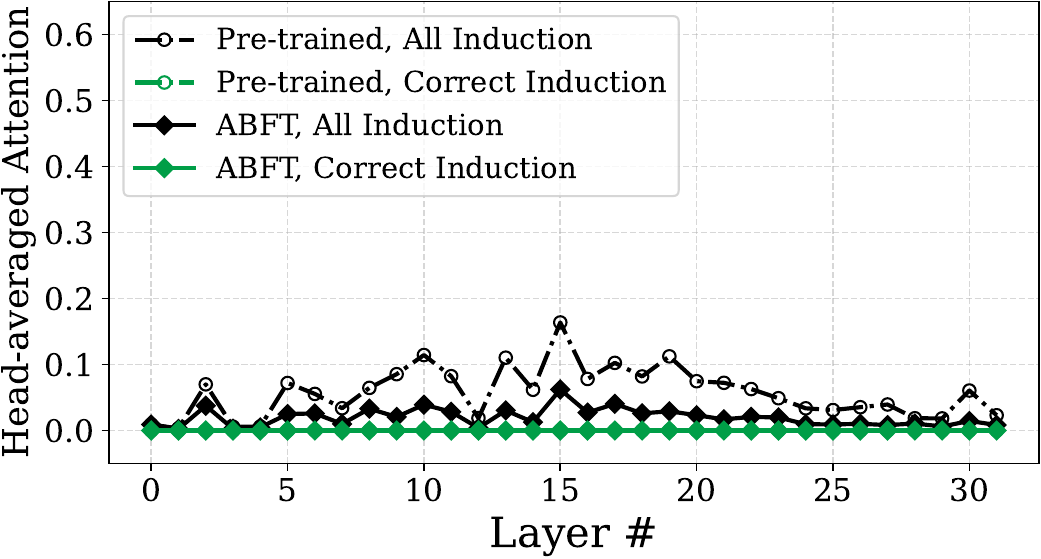}
    \vspace{-1.5\baselineskip}
    \caption{Visualization of induction attention scores on unseen label settings (Llama3 8B, TREC).}
    \label{fig:unseen_attn_vis}
    \vspace{-1\baselineskip}
\end{figure}

First, we train a Llama3 8B on TREC (a 6-way classification dataset) with the ABFT method. Then, to test the trained ABFT model on the unseen label condition, we build special test inputs: for each query with label $l^*$, we choose $k=4$ demonstrations with label $l\not=l^*$, and utilize the standard template shown in Table~\ref{tab:prompt_templates} to build the inputs, then test the accuracy. Notice that during the training, no special sampling for the inputs is conducted, i.e., the training is not under the unseen label setting, so that such an experiment protocol also confirms the generalization of ABFT methods on a different distribution. Moreover, we repeat the induction head visualization shown in Fig.~\ref{fig:attention_score} on the unseen label condition in Fig.~\ref{fig:unseen_attn_vis}, where the induction heads in the ABFT model are almost fully suppressed but with considerable inference accuracy, which implies a new inference mechanism.

\subsection{Experiment Protocol of Stability against Prompting}
\label{appendix:stability}

In Table~\ref{table:robustness}, we test whether the prediction of ICL is stable against various \textbf{(1)} prompt templates and \textbf{(2)} demonstration sampling, on the \textsc{StaICC-Diag}~\cite{cho2025staicc} benchmark, whose method is described briefly below.

\paragraph{Method.} To test the prediction robustness against prompt templates / demonstration sampling, we repeat several predictions for each query on various prompt templates / demonstration sampling, and calculate the ratio of the maximum consistent group (e.g., we get 6 positive and 3 negative predictions on one query, then the ratio is $\max(6,3)/(6+3)=2/3$). The robustness metrics are the average value of the whole dataset. Refer to~\citet{cho2025staicc} for the detailed implementation. Notice that only the consistency is tested in these experiments, without observing the accuracy.

\paragraph{Result.} The robustness metrics among prompt templates / demonstration sampling averaged on all 8 datasets before and after ABFT are shown in Table~\ref{table:robustness}, where both terms of the robustness are significantly improved after ABFT, suggesting that ABFT stabilizes ICL for various contexts, providing higher efficiency on prompt designing. Also, given the results with mitigating prediction sensitivity and bias against prompt templates / demonstration sampling, which is consistent with the objective of output calibration~\cite{zhao2021calibrate, fei2023mitigating, han2023prototypical, zhou2024batch, jiang-etal-2023-generative, cho-etal-2025-token}, ABFT can be regarded as an implicit calibration inside the LLM. 



\section{Parameter Shift after ABFT against Layers}
\label{appendix:parameter_shift}

\begin{figure*}[t]
    \centering
    \includegraphics[width=0.7\linewidth]{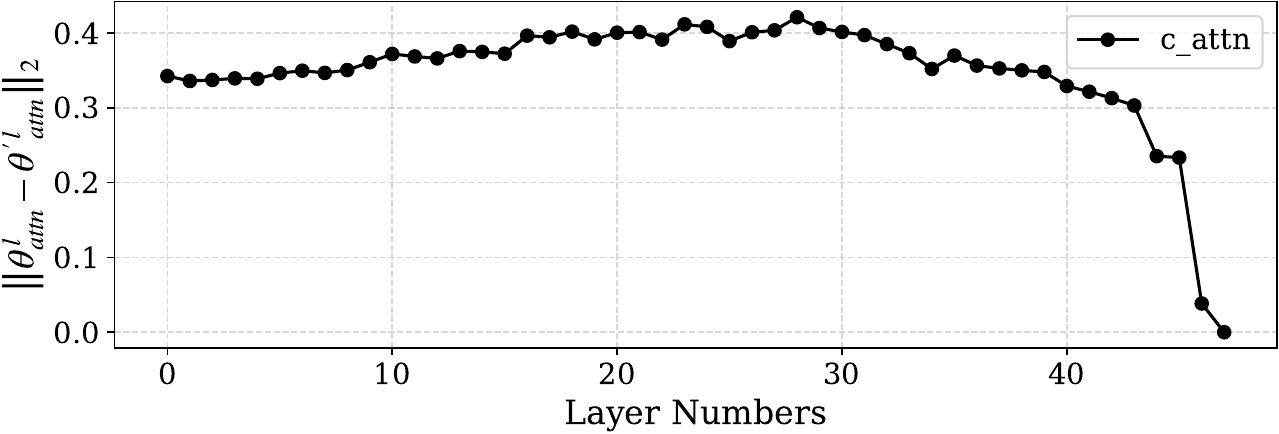}
    \caption{Shifting distance before and after ABFT on the \texttt{c\_attn} matrix of GPT2-XL and SST2.}
    \label{fig:norm_GPT2}
\end{figure*}

\begin{figure*}[t]
    \centering
    \includegraphics[width=0.7\linewidth]{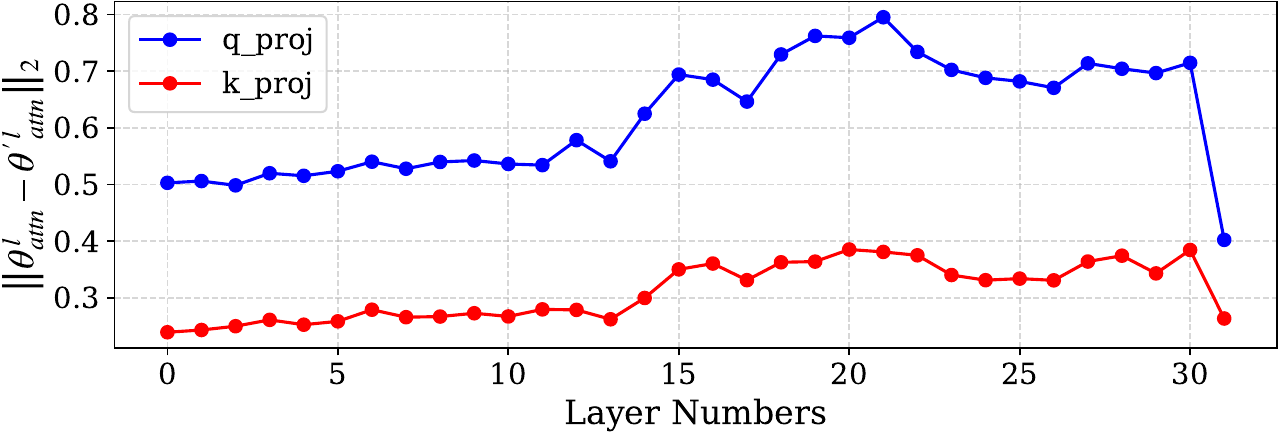}
    \caption{Shifting distance before and after ABFT on the \texttt{q\_proj} and \texttt{k\_proj} matrix of Llama3-8B and SST2.}
    \label{fig:norm_llama3_8B}
\end{figure*}

\begin{figure*}[t]
    \centering
    \includegraphics[width=0.7\linewidth]{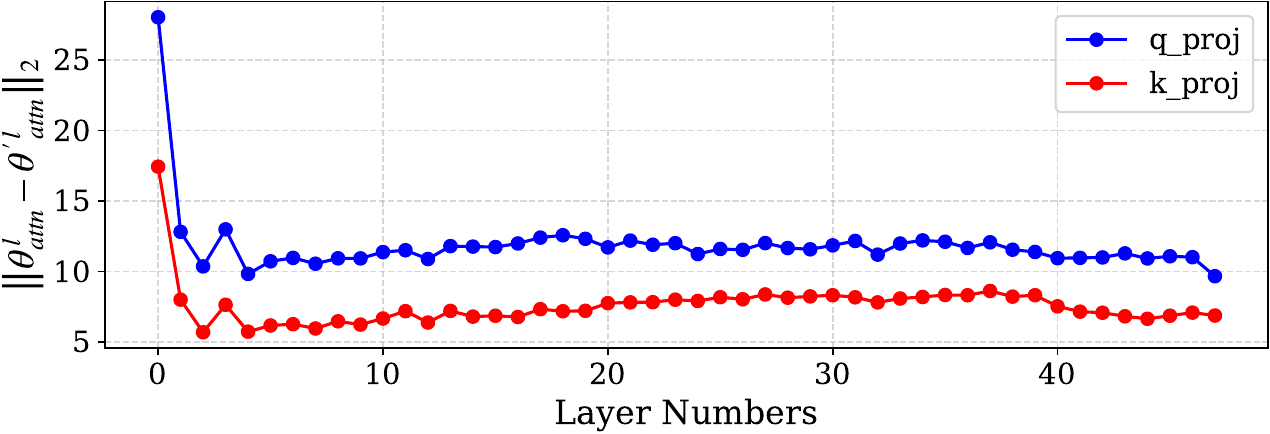}
    \caption{Shifting distance before and after ABFT on the \texttt{q\_proj} and \texttt{k\_proj} matrix of Llama3-42B and SST2.}
    \label{fig:norm_llama3_42B}
\end{figure*}

We utilize the Frobenius norm to visualize the shifting distance of the parameter matrix $\theta$ before and after ABFT ($\theta'$) as $\Vert\theta-\theta'\Vert_2$. The results are shown in Fig.~\ref{fig:norm_GPT2},~\ref{fig:norm_llama3_8B},~\ref{fig:norm_llama3_42B}, where, although each model exhibits its own pattern in terms of distance across layer numbers, certain layers consistently show significantly lower distances within every model. 

Moreover, even though the early layers accumulate more gradients (since the gradients from each later layer propagate backward to them), the peak of the shifting distance typically appears in the middle to later layers. This observation is consistent with previous works on Induction Heads~\cite{cho2025revisiting}.

\section{Augmentation Results}

\subsection{Augmentation Results for Data Efficiency (Fig.~\ref{fig:data_eff})}
\label{appendix:more_res_on_DE}

\begin{figure}[t]
    \centering
    \includegraphics[width=0.9\linewidth]{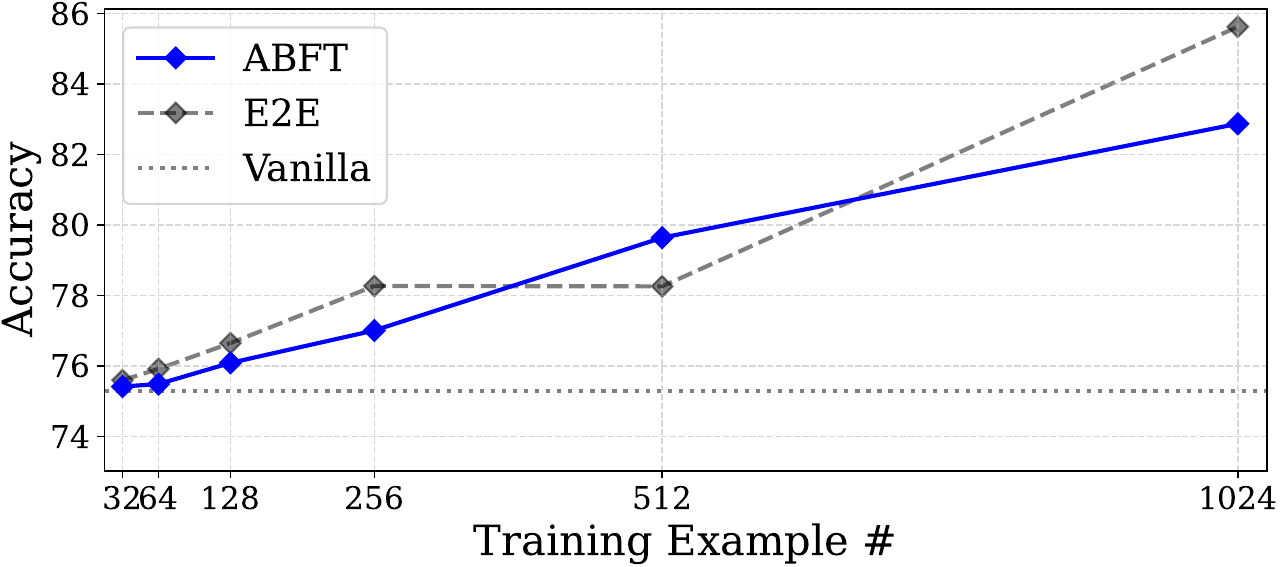}
    \vspace{-0.6\baselineskip}
    \caption{Accuracy against training set size as a metric of data efficiency, for Qwen2.5 32B.}
    \label{fig:dataeff_32B}
\end{figure}

We repeat the data efficiency experiments shown in Fig.~\ref{fig:data_eff} on Qwen2.5 32B, as shown in Fig.~\ref{fig:dataeff_32B}. The results are globally consistent with Fig.~\ref{fig:data_eff}.

\subsection{Augmentation Results for Number of Induction Heads against Training Processing (Fig.~\ref{fig:induction_head_number})}
\label{appendix:attn_head_n_more}

\begin{figure*}[t]
    \centering
    \subfloat[Llama3-8B SST2 (Fig.~\ref{fig:induction_head_number})]{\includegraphics[width=0.47\linewidth]{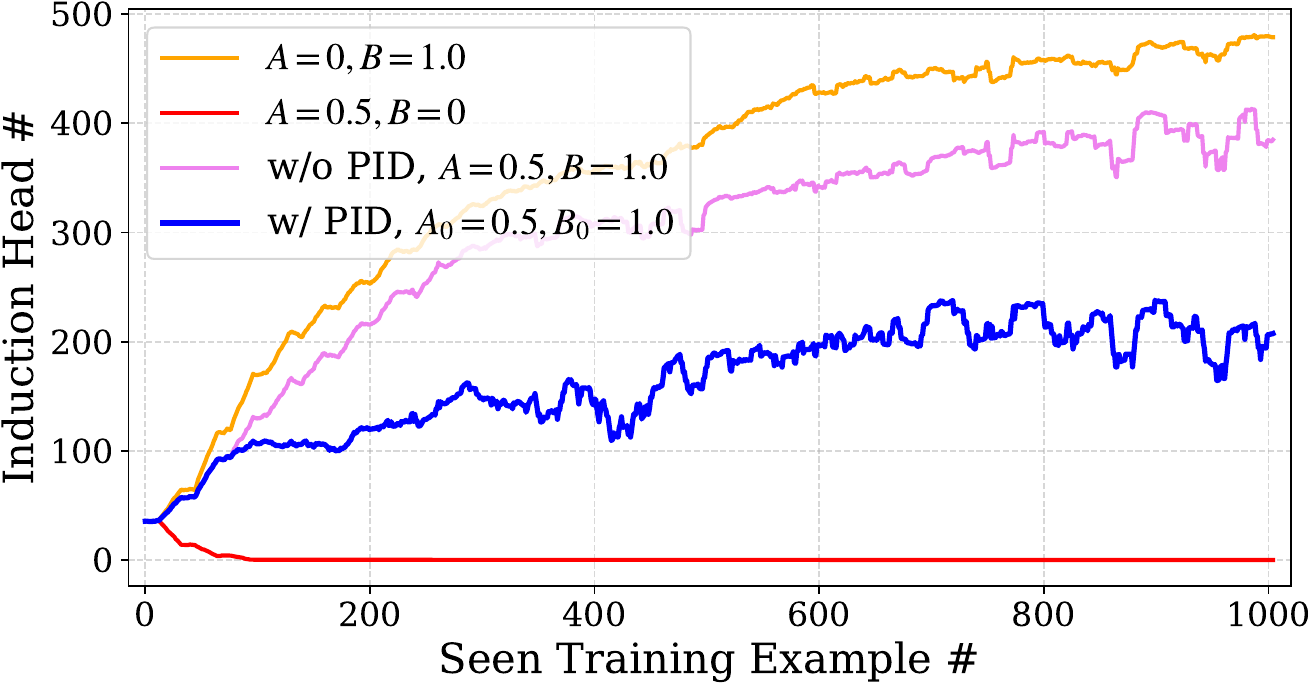}}\hfill
    \subfloat[Llama3-8B MR]{\includegraphics[width=0.47\linewidth]{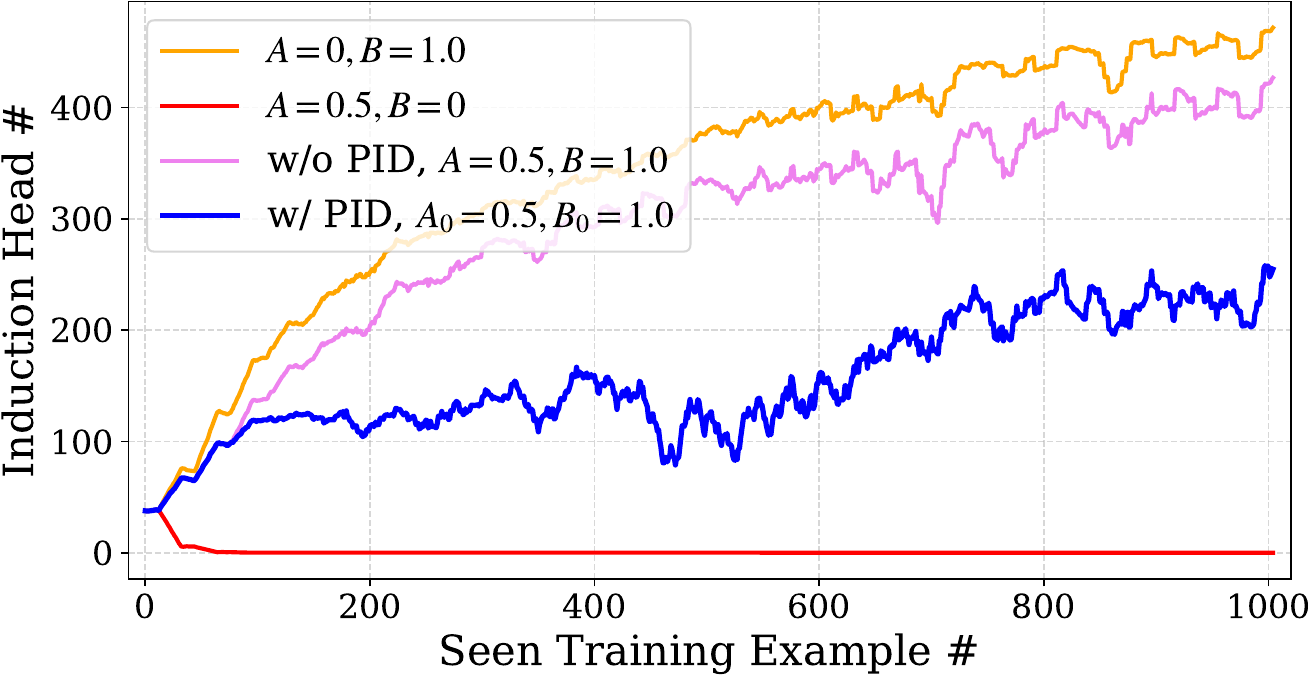}}\\
    \subfloat[Llama3-8B FP]{\includegraphics[width=0.47\linewidth]{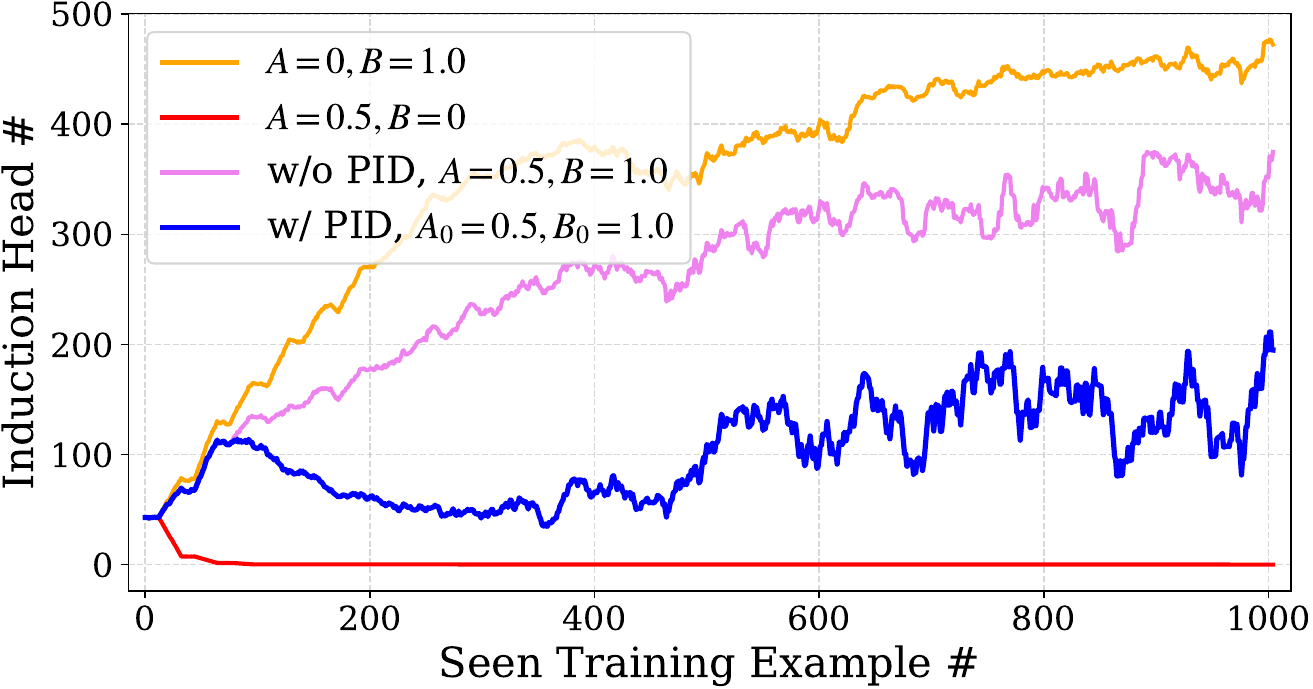}}\hfill
    \subfloat[Llama3-8B SST5]{\includegraphics[width=0.47\linewidth]{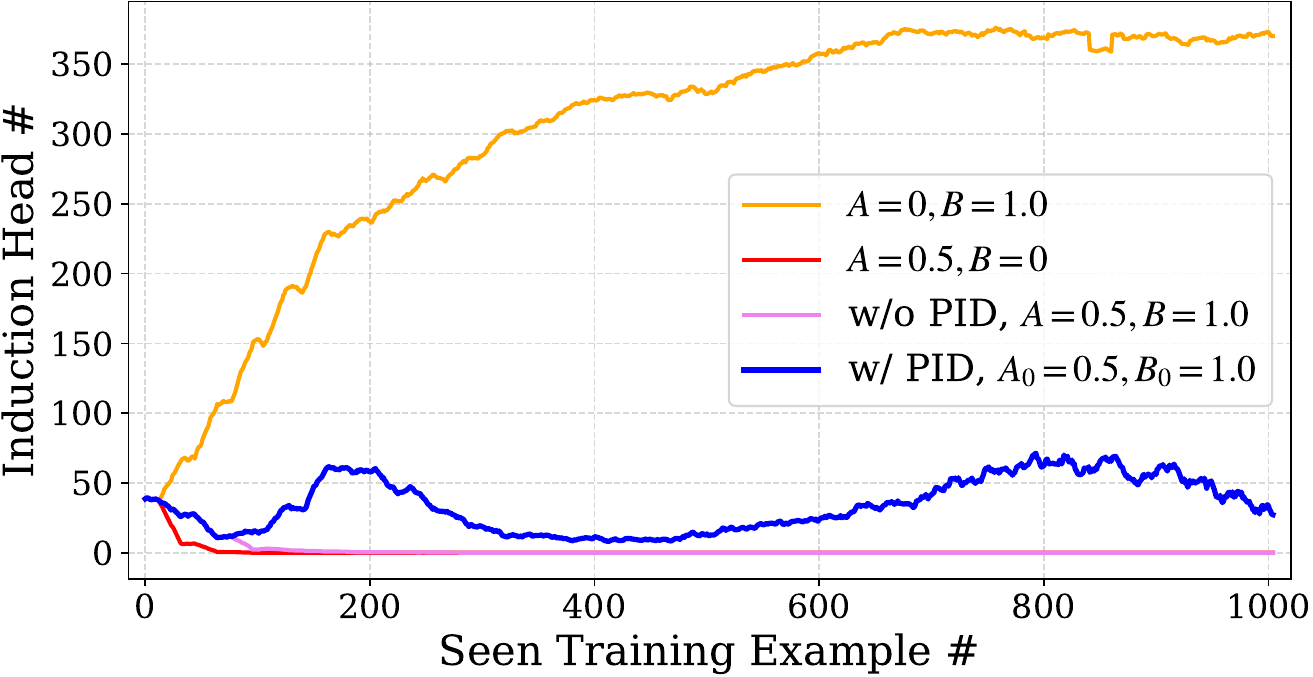}}\\
    \subfloat[Llama3-8B TREC]{\includegraphics[width=0.47\linewidth]{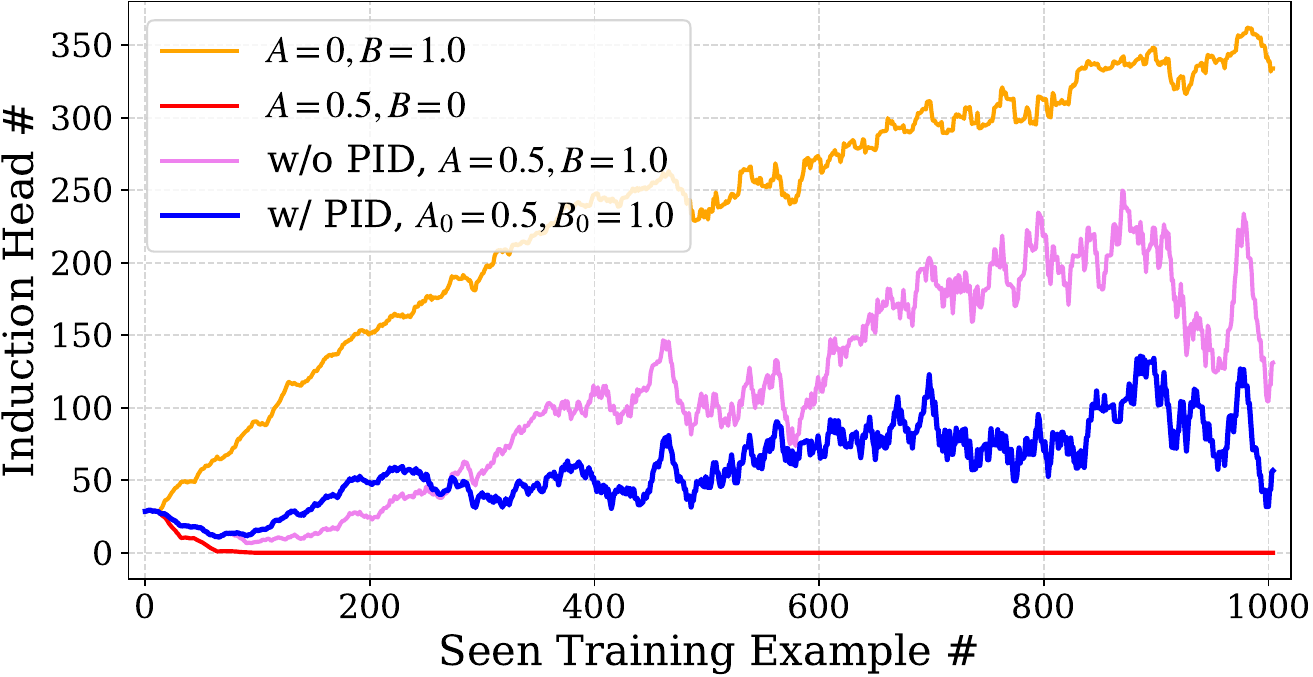}}\hfill
    \subfloat[Llama3-8B SUBJ]{\includegraphics[width=0.47\linewidth]{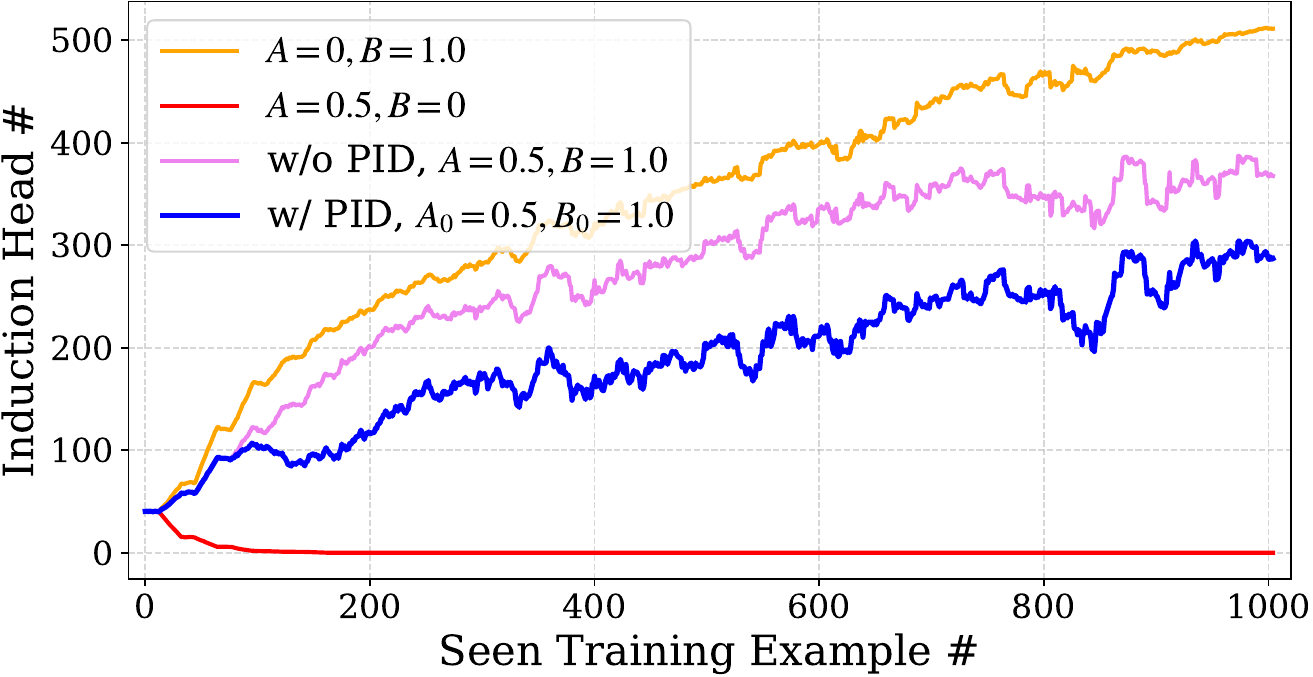}}\\
    \subfloat[Llama3-8B TEE]{\includegraphics[width=0.47\linewidth]{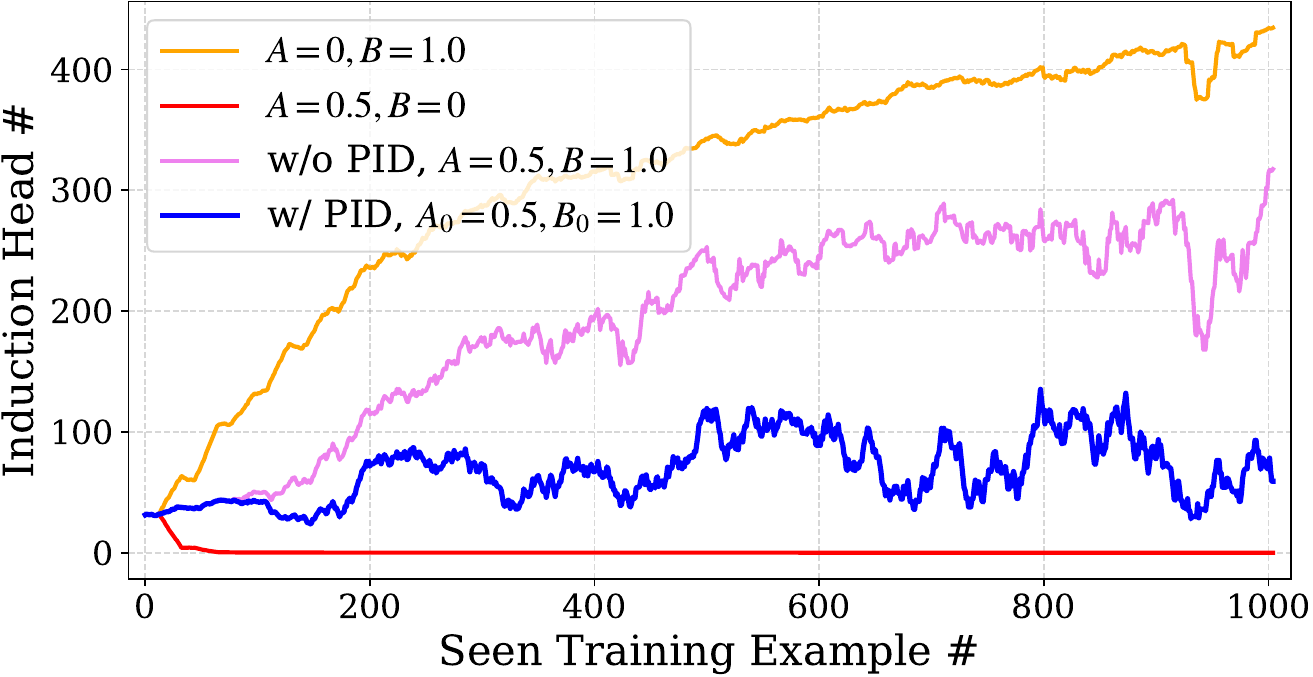}}\hfill
    \subfloat[Llama3-8B TEH]{\includegraphics[width=0.47\linewidth]{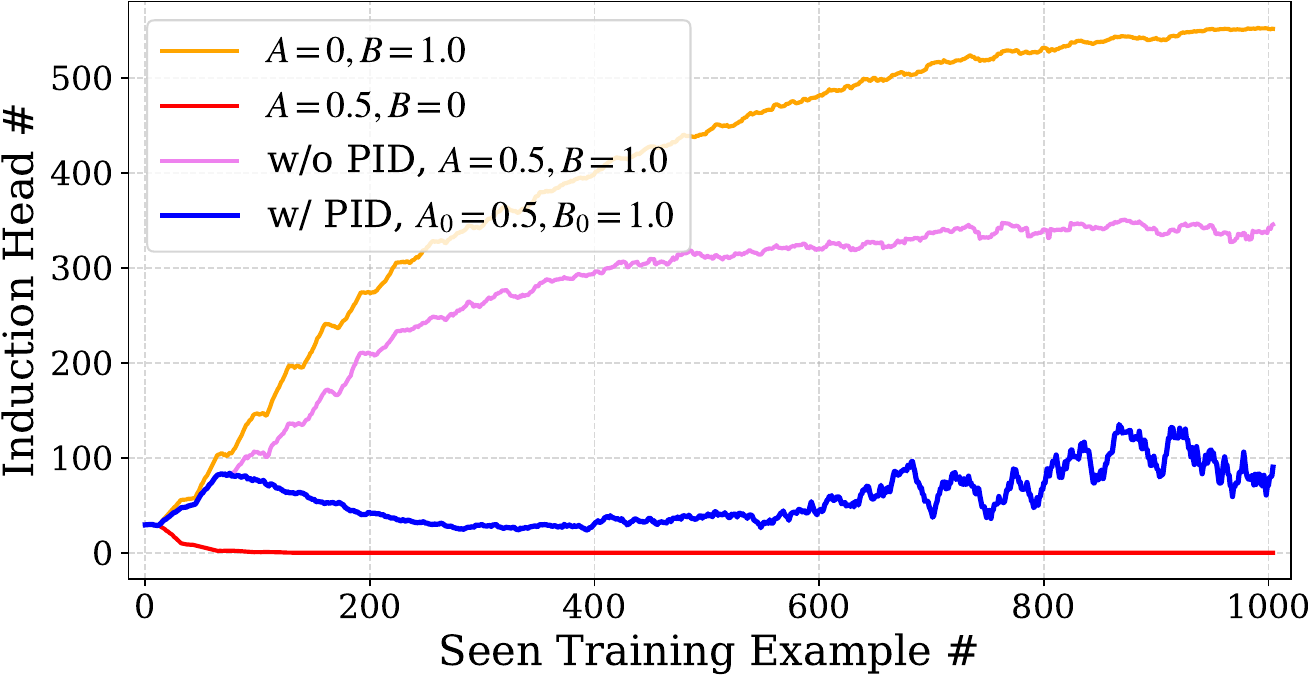}}\\
    \caption{Induction head numbers along training dynamics on Llama3-8B and all 8 datasets.}
    \label{fig:attention_head_n_more}
\end{figure*}

\begin{figure*}[t]
    \centering
    \subfloat[Falcon3-7B SST2]{\includegraphics[width=0.47\linewidth]{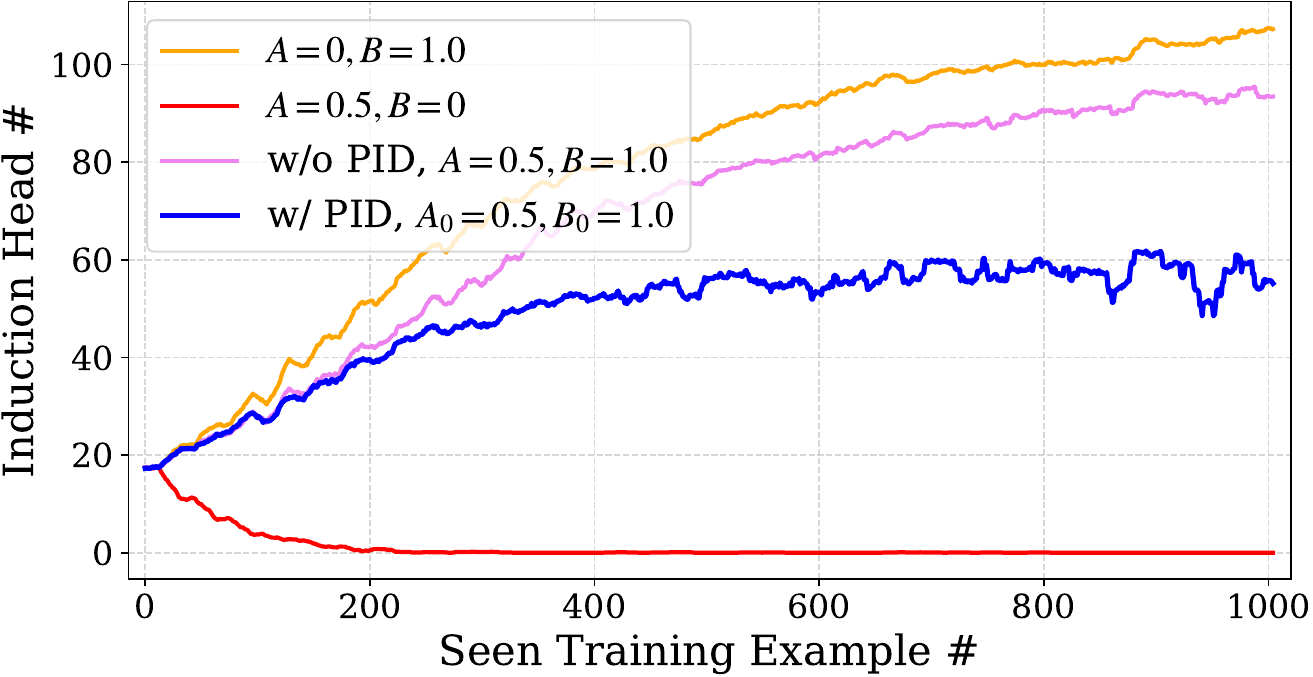}}\hfill
    \subfloat[Falcon3-7B MR]{\includegraphics[width=0.47\linewidth]{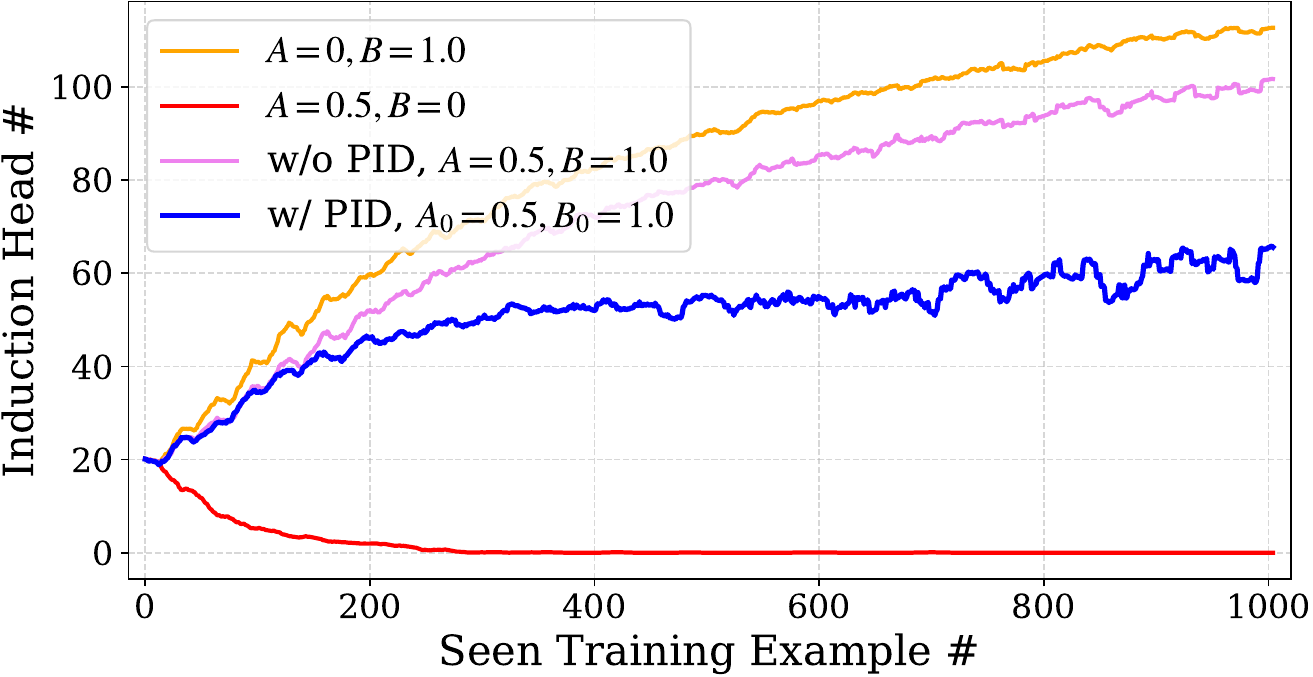}}\\
    \subfloat[Falcon3-7B FP]{\includegraphics[width=0.47\linewidth]{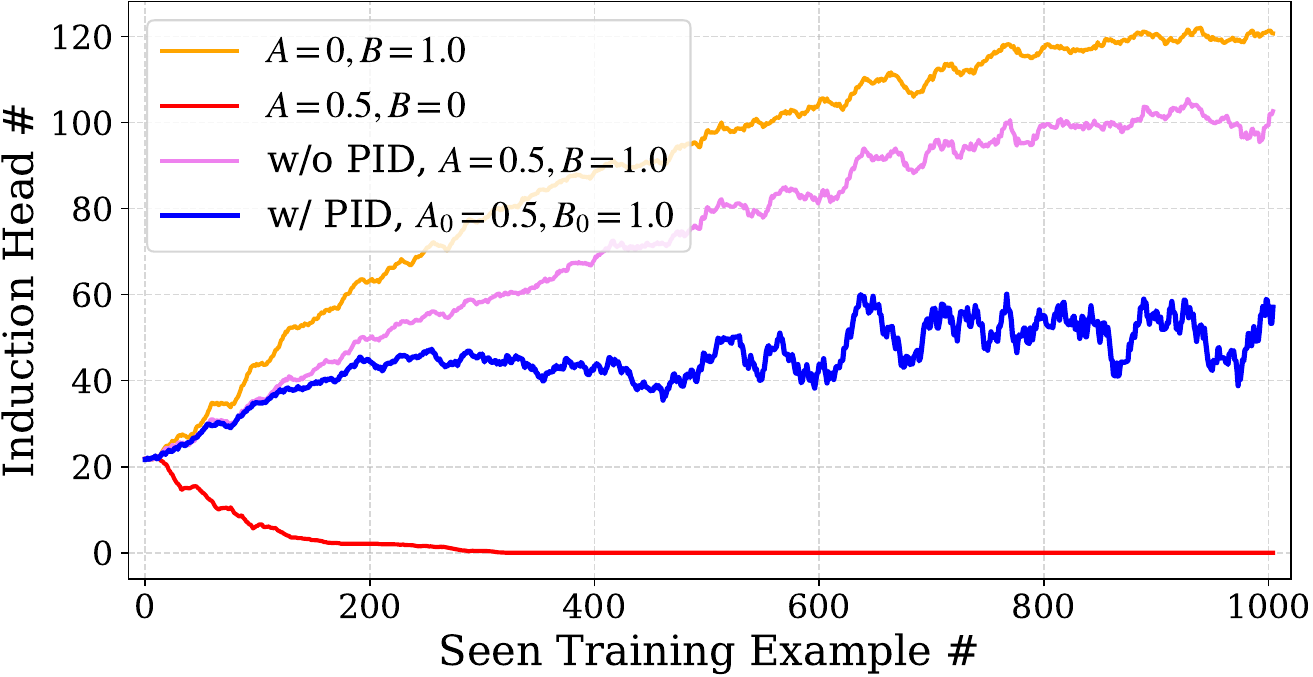}}\hfill
    \subfloat[Falcon3-7B SST5]{\includegraphics[width=0.47\linewidth]{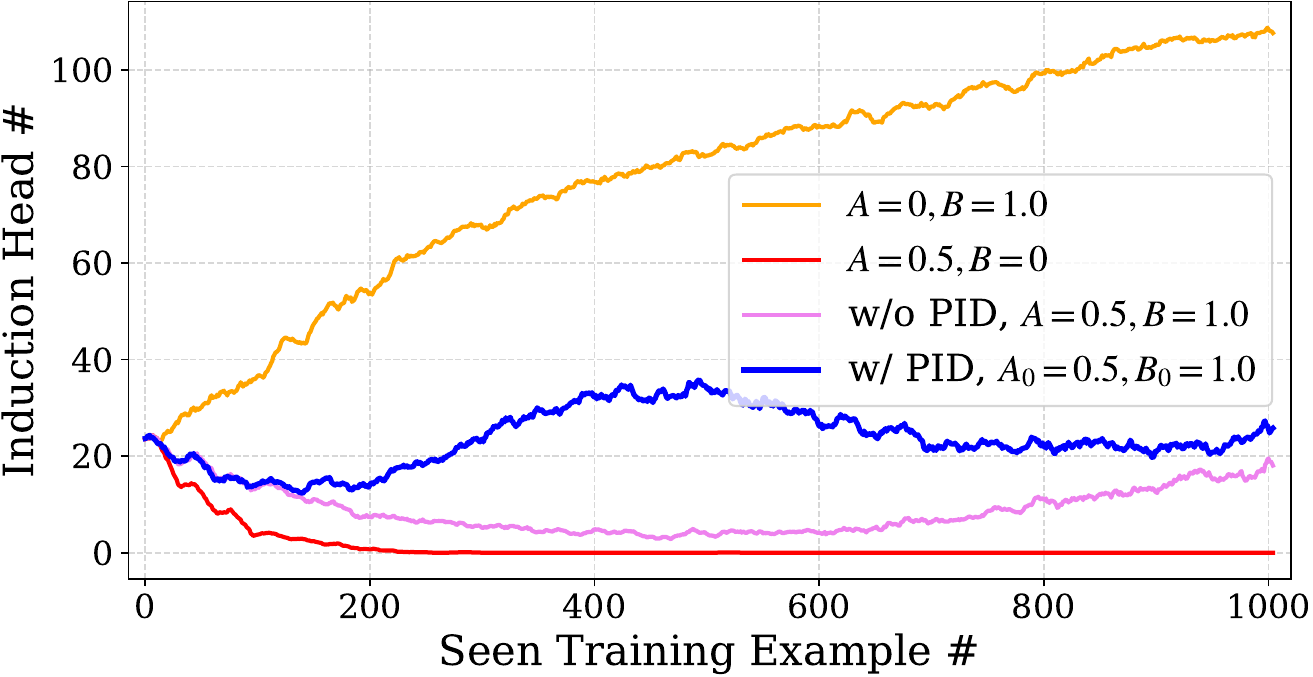}}\\
    \subfloat[Falcon3-7B TREC]{\includegraphics[width=0.47\linewidth]{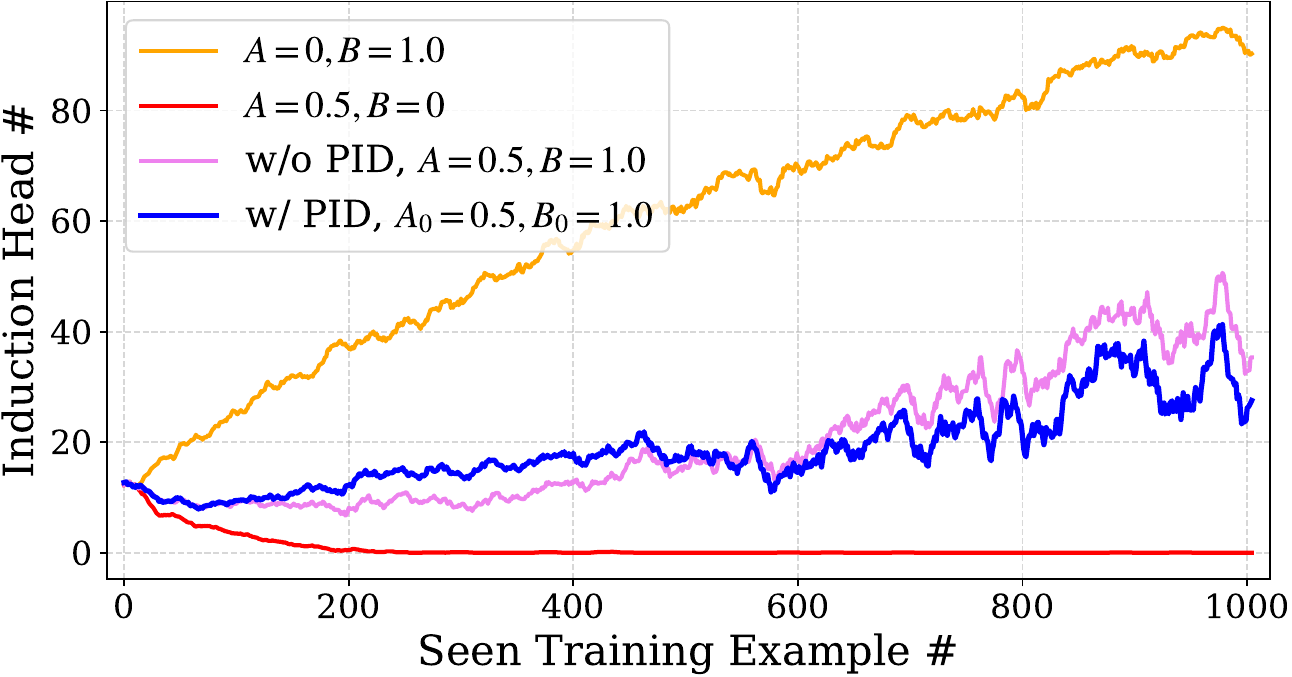}}\hfill
    \subfloat[Falcon3-7B SUBJ]{\includegraphics[width=0.47\linewidth]{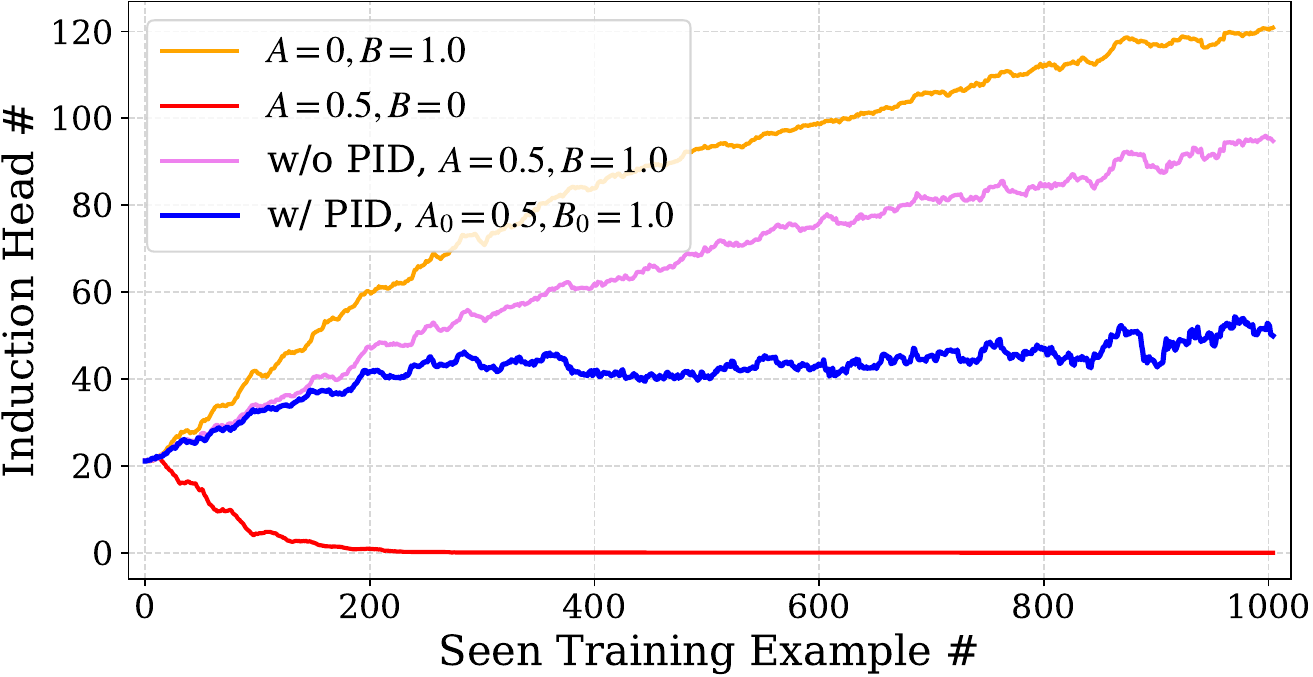}}\\
    \subfloat[Falcon3-7B TEE]{\includegraphics[width=0.47\linewidth]{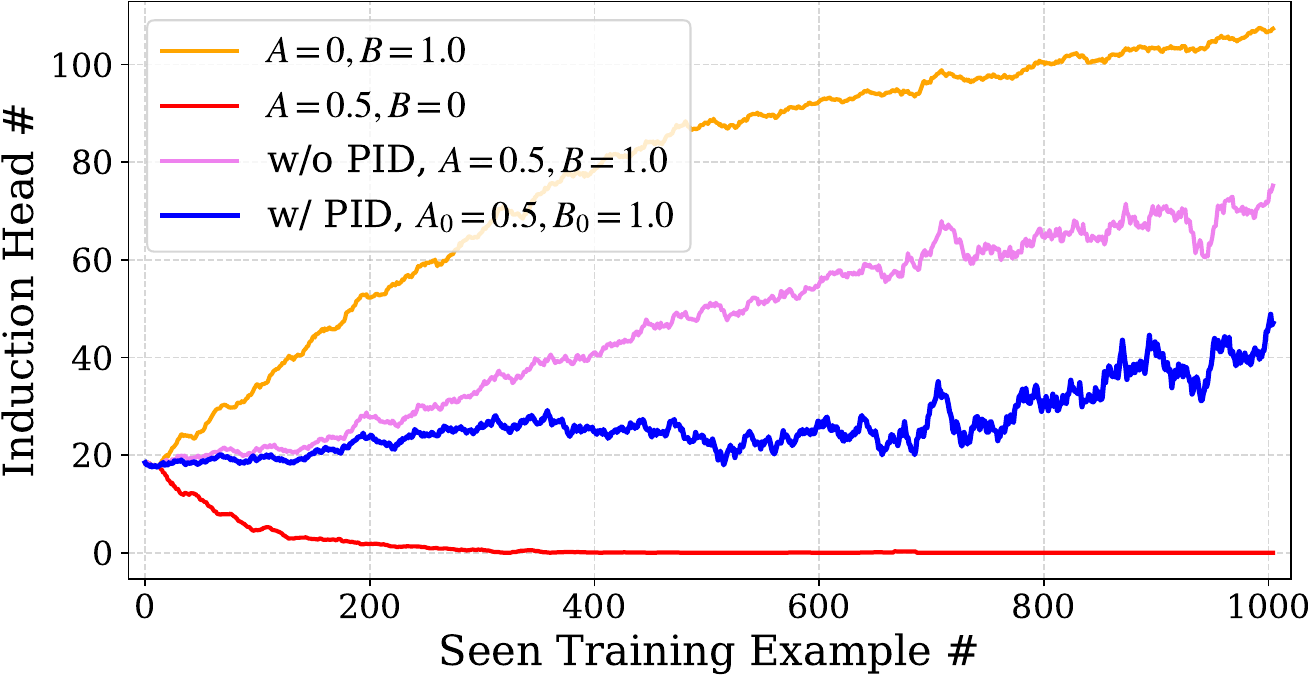}}\hfill
    \subfloat[Falcon3-7B TEH]{\includegraphics[width=0.47\linewidth]{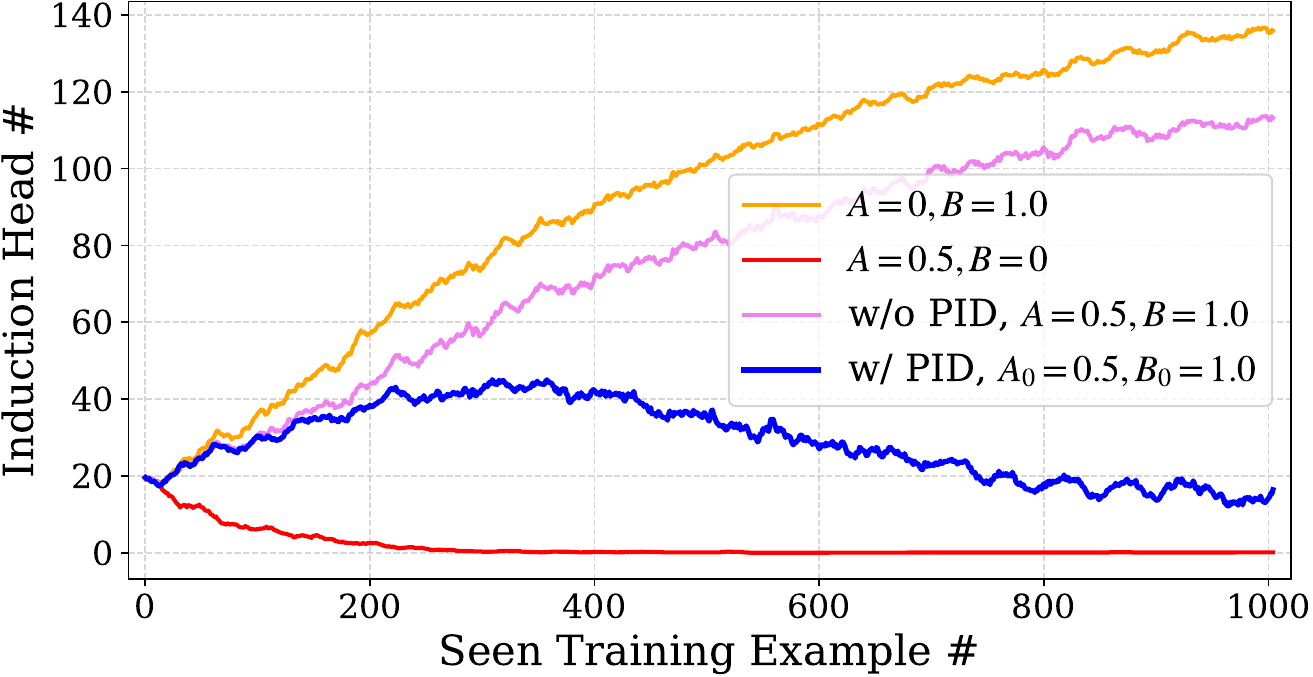}}\\
    \caption{Induction head numbers along training dynamics on Falcon3-7B and all 8 datasets.}
    \label{fig:attention_head_n_more_2}
\end{figure*}

We repeat the visualization of the number of induction heads against the training processing under various settings on Llama3 8B and Falcon3 7B as shown in Fig.~\ref{fig:attention_head_n_more} and~\ref{fig:attention_head_n_more_2}. The results are globally consistent with Fig.~\ref{fig:induction_head_number}.

\begin{figure*}[t]
    \centering
    \includegraphics[width=\linewidth]{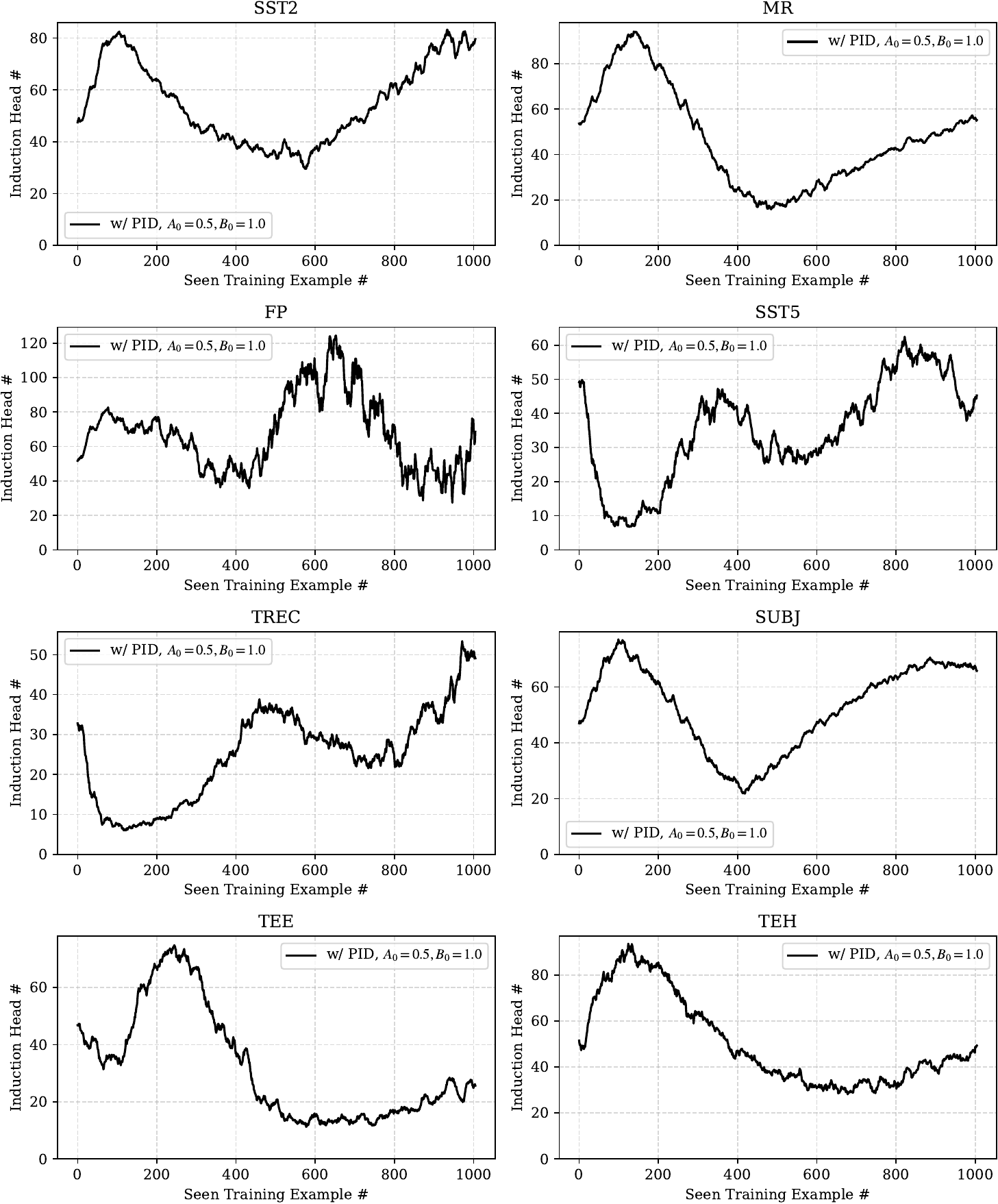}
    \caption{Induction head numbers along training dynamics on GPT2-Large and all 8 datasets.}
    \label{appendix.induction_n_gpt2l}
\end{figure*}

\begin{figure*}[t]
    \centering
    \includegraphics[width=\linewidth]{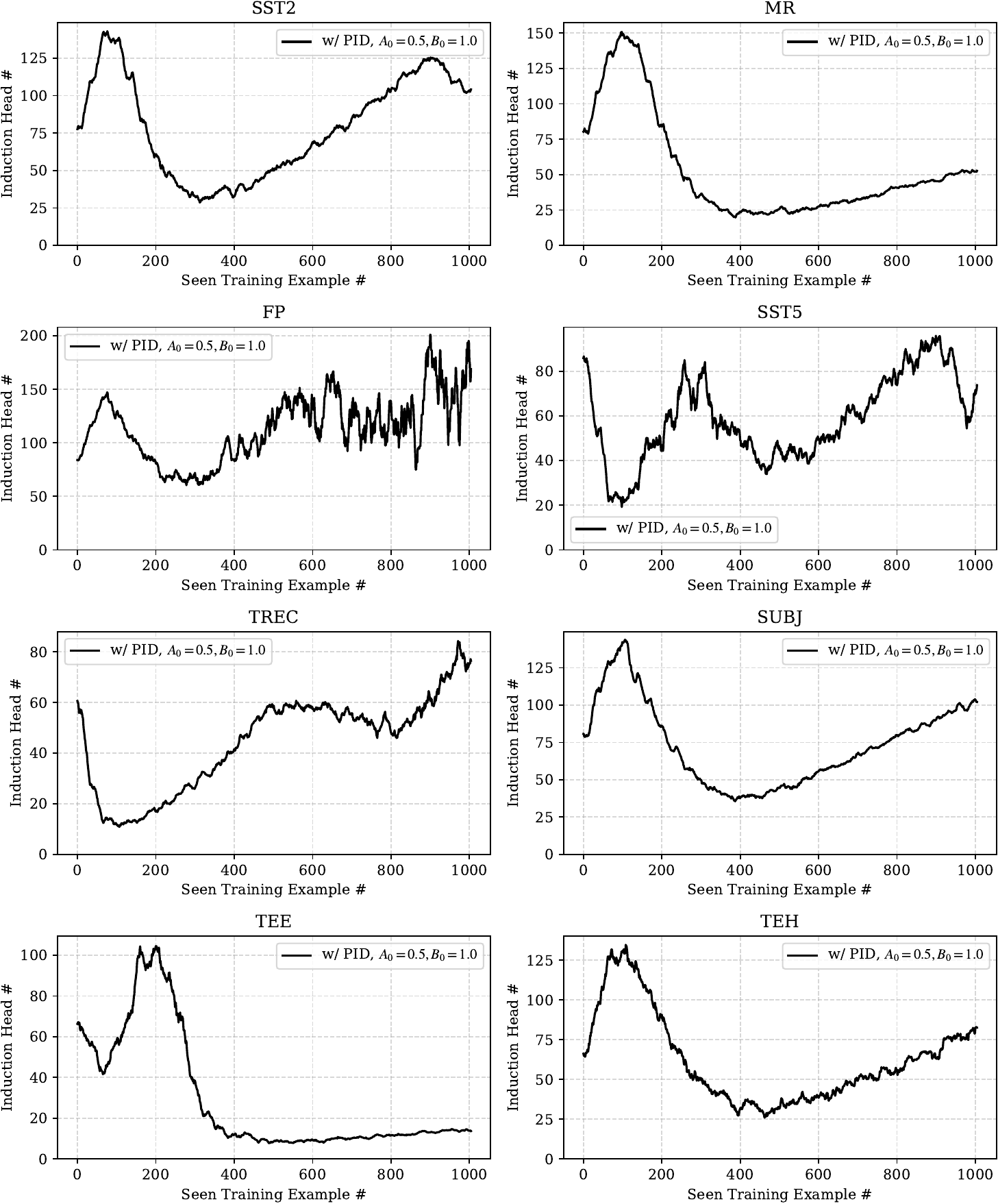}
    \caption{Induction head numbers along training dynamics on GPT2-XL and all 8 datasets.}
    \label{appendix.induction_n_gpt2xl}
\end{figure*}

\begin{figure*}[t]
    \centering
    \includegraphics[width=\linewidth]{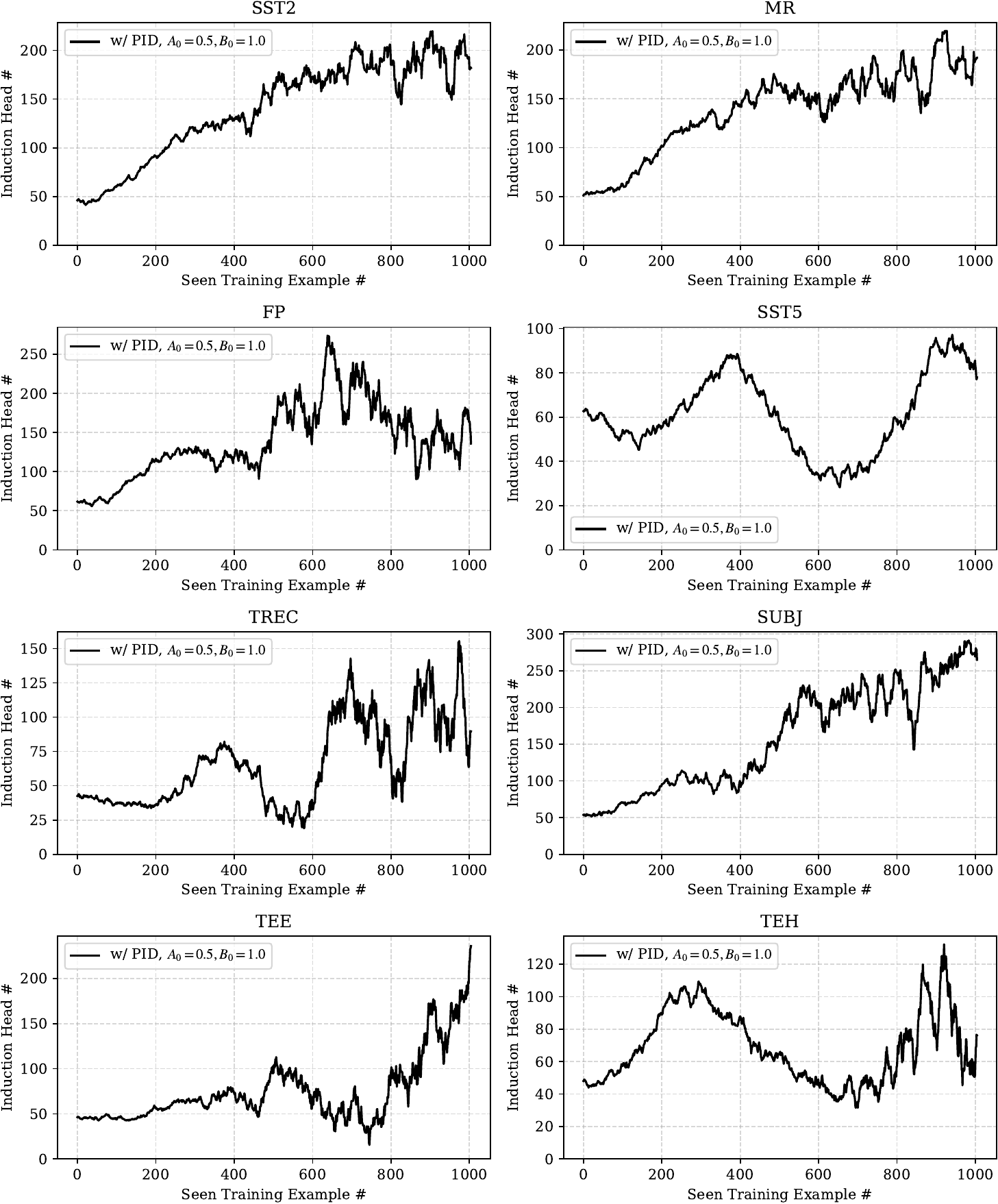}
    \caption{Induction head numbers along training dynamics on DeepSeek-R1 and all 8 datasets.}
    \label{appendix.induction_n_r1}
\end{figure*}

\begin{figure*}[t]
    \centering
    \includegraphics[width=\linewidth]{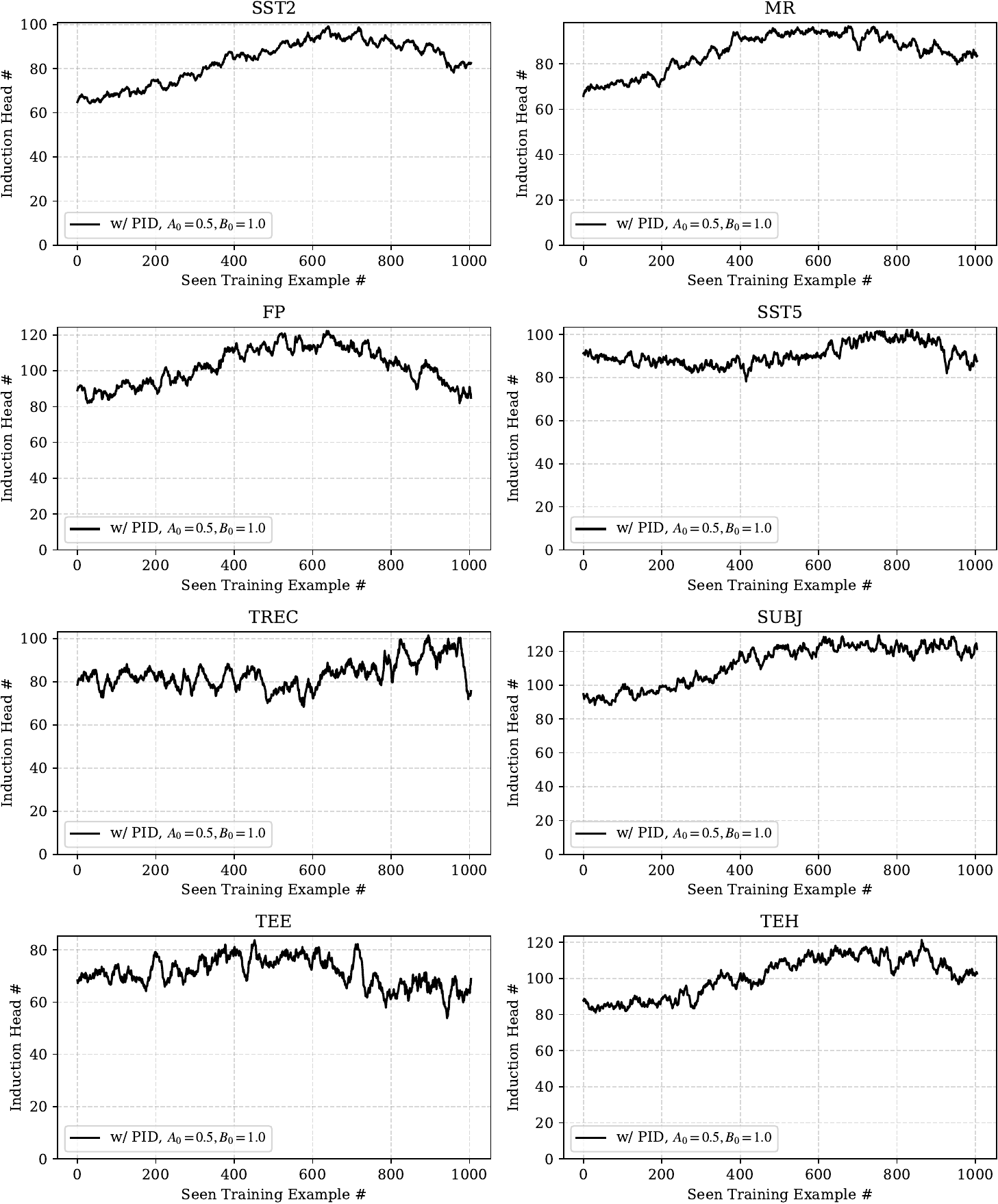}
    \caption{Induction head numbers along training dynamics on Qwen2.5-32B and all 8 datasets.}
    \label{appendix.induction_n_qwen2.5}
\end{figure*}

\begin{figure*}[t]
    \centering
    \includegraphics[width=\linewidth]{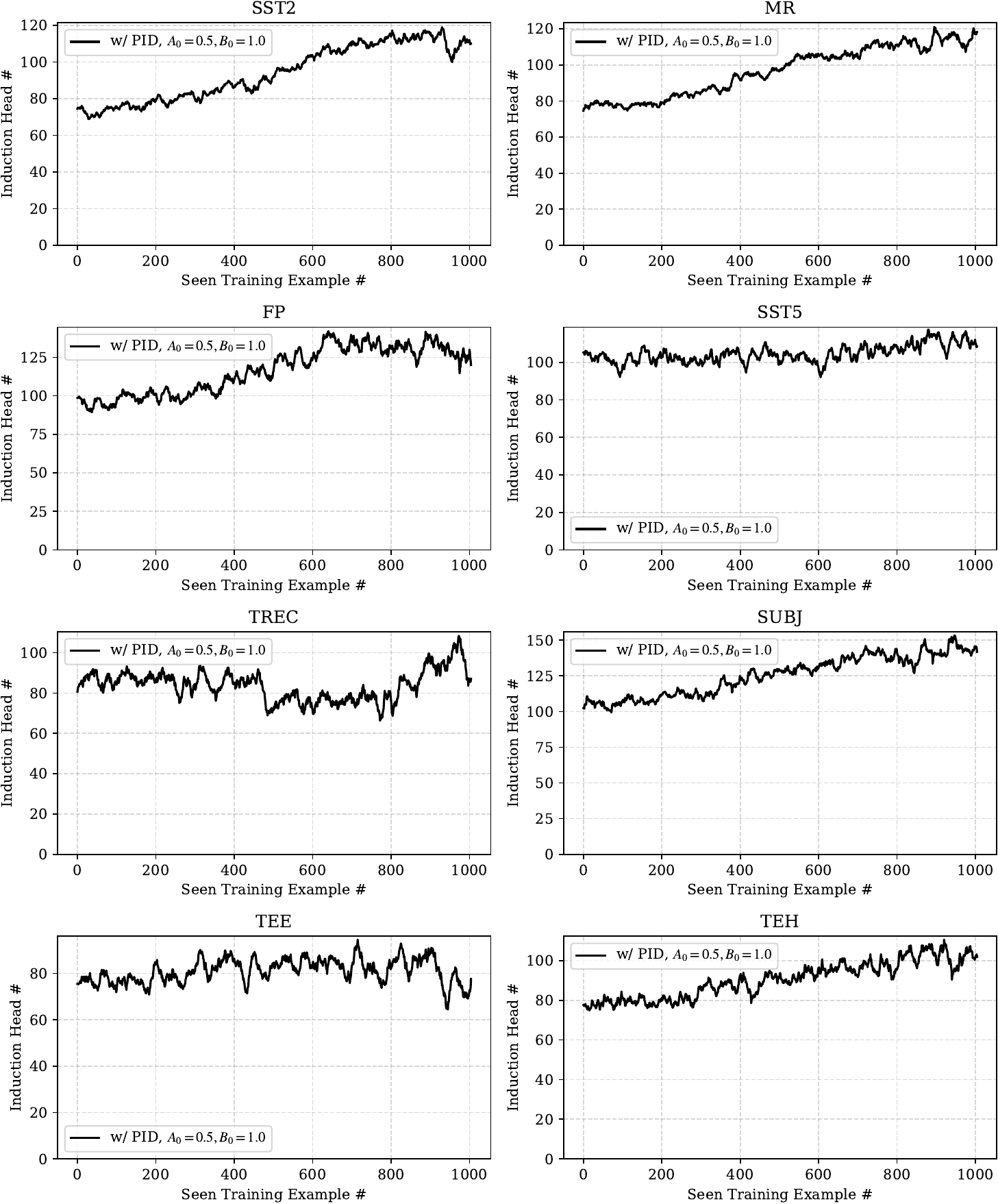}
    \caption{Induction head numbers along training dynamics on SimpleScaling s1.1 and all 8 datasets.}
    \label{appendix.induction_n_s1}
\end{figure*}

\begin{figure*}[t]
    \centering
    \includegraphics[width=\linewidth]{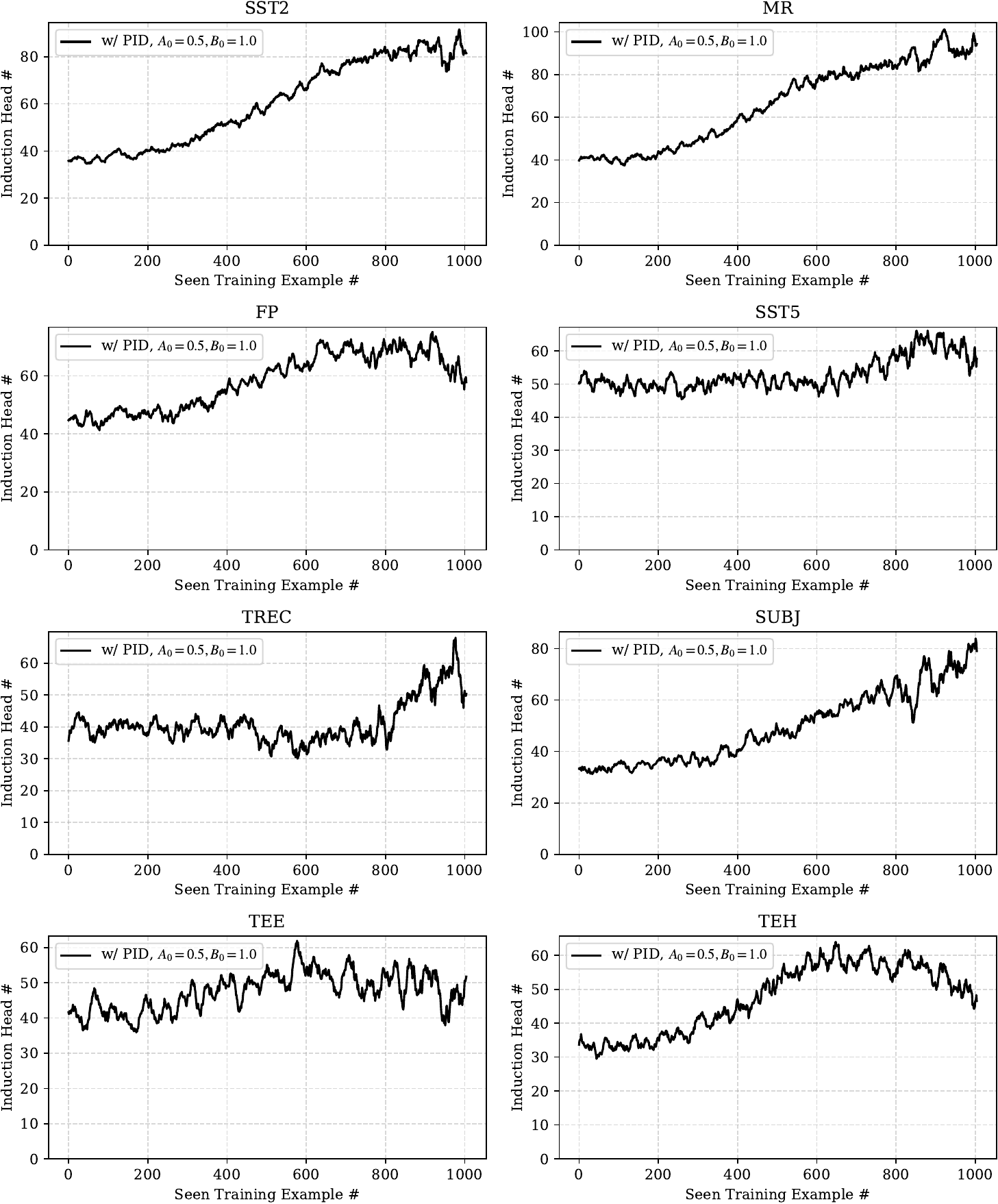}
    \caption{Induction head numbers along training dynamics on Llama3-42B and all 8 datasets.}
    \label{appendix.induction_llama3_42B}
\end{figure*}

\begin{figure*}[t]
    \centering
    \includegraphics[width=\linewidth]{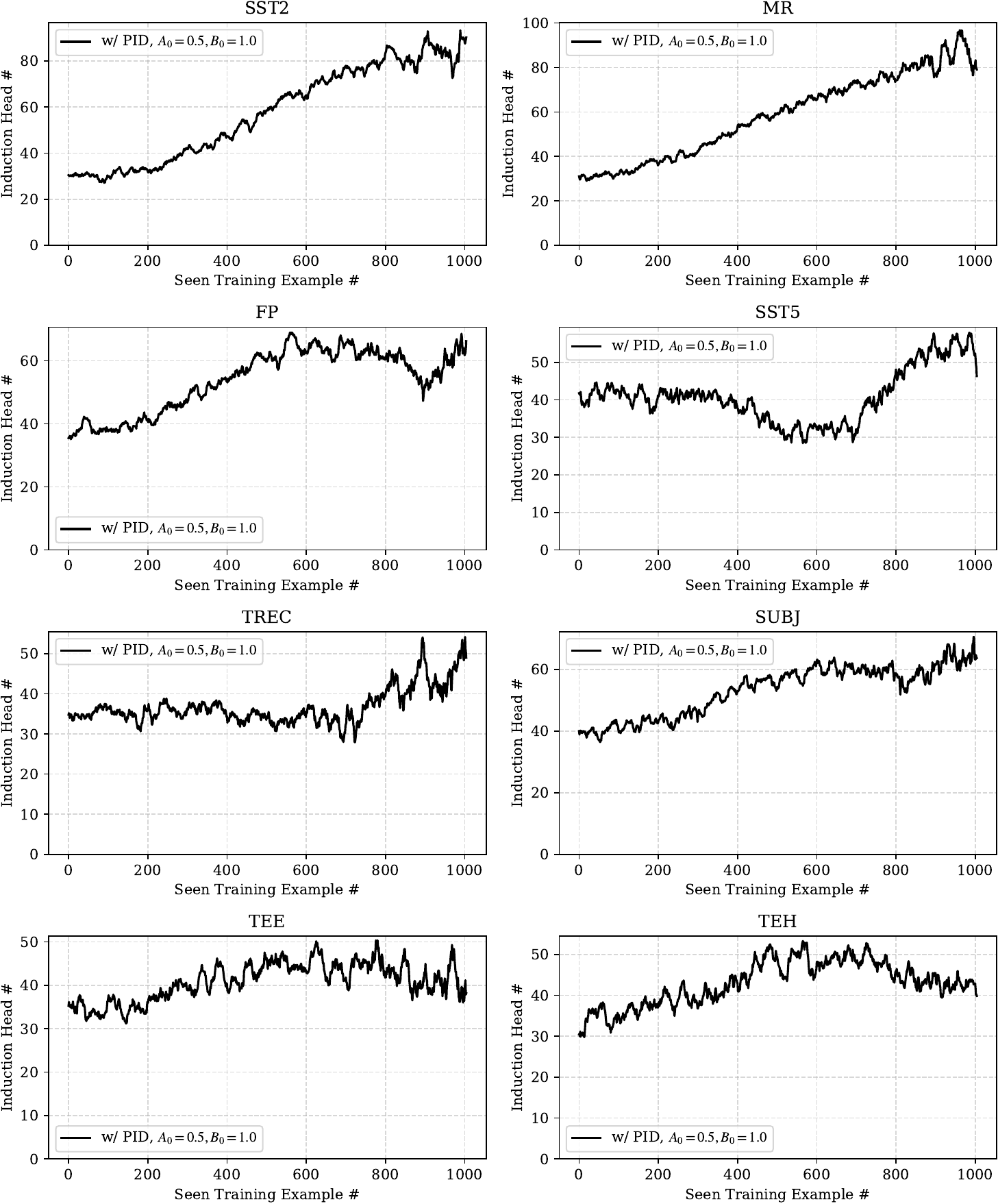}
    \caption{Induction head numbers along training dynamics on Llama3-56B and all 8 datasets.}
    \label{appendix.induction_llama3_56B}
\end{figure*}

Moreover, we visualize the number of induction heads on only standard settings, as shown in Fig.~\ref{appendix.induction_n_gpt2l}-\ref{appendix.induction_llama3_56B} for reference.

\subsection{Augmentation Results for Attention Visualization (Fig.~\ref{fig:attention_score} and~\ref{fig:attention_visualization_3})}
\label{appendix:attn_vis_more}

\begin{figure}[t]
    \centering
    \includegraphics[width=0.9\linewidth]{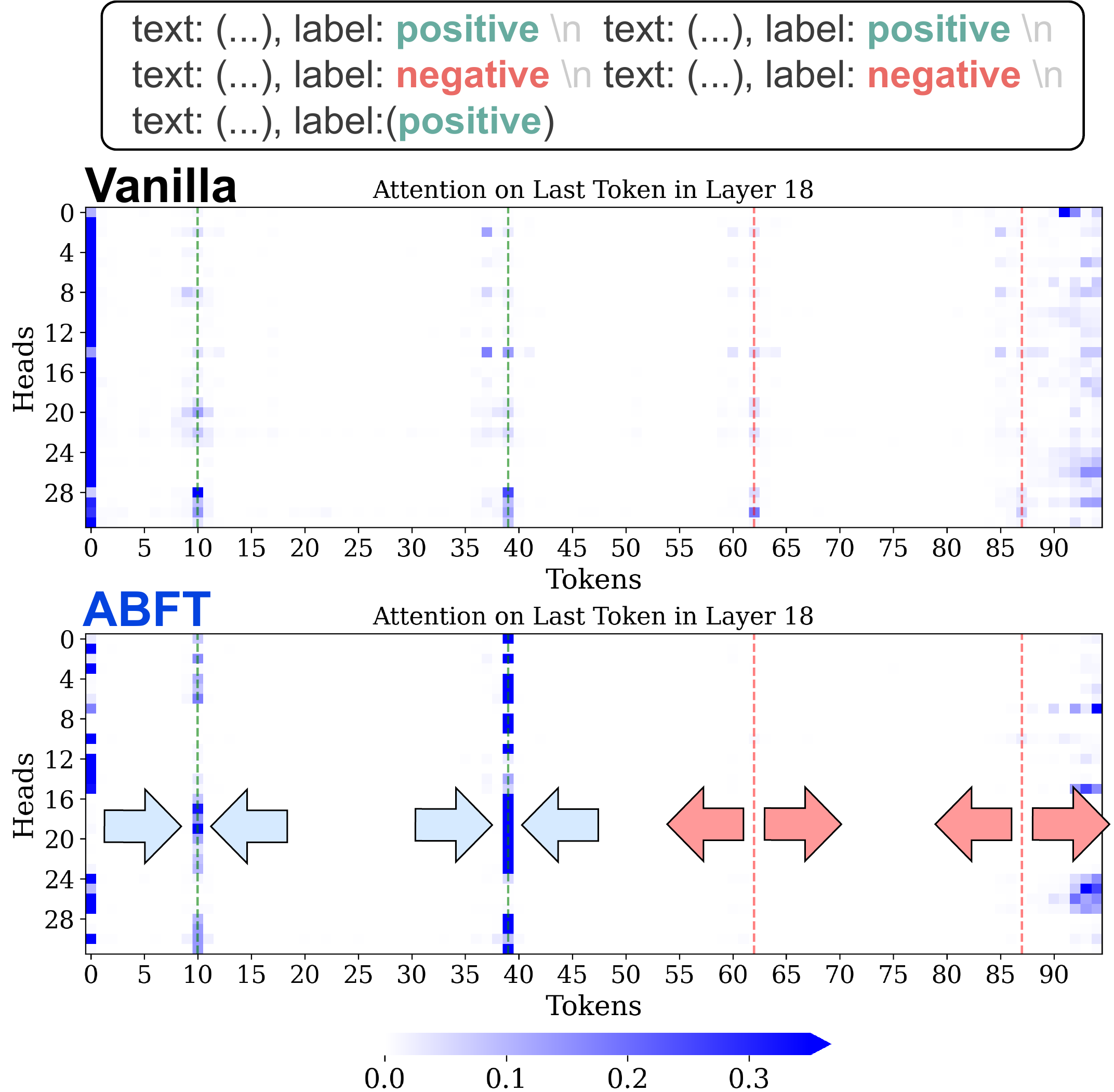}
    \vspace{-0.6\baselineskip}
    \caption{Attention score visualization on the last token of ICL input of every attention head (vertical axis) towards each token. Label tokens and their contents are marked with dotted lines. Refer to Appendix~\ref{appendix:attn_vis_more} for more examples and layers. \textbf{ABFT successfully focuses attention scores to correct labels}.}
    \label{fig:attention_visual}
\end{figure}

\begin{sidewaysfigure*}[t]
    \centering
    \includegraphics[width=\linewidth]{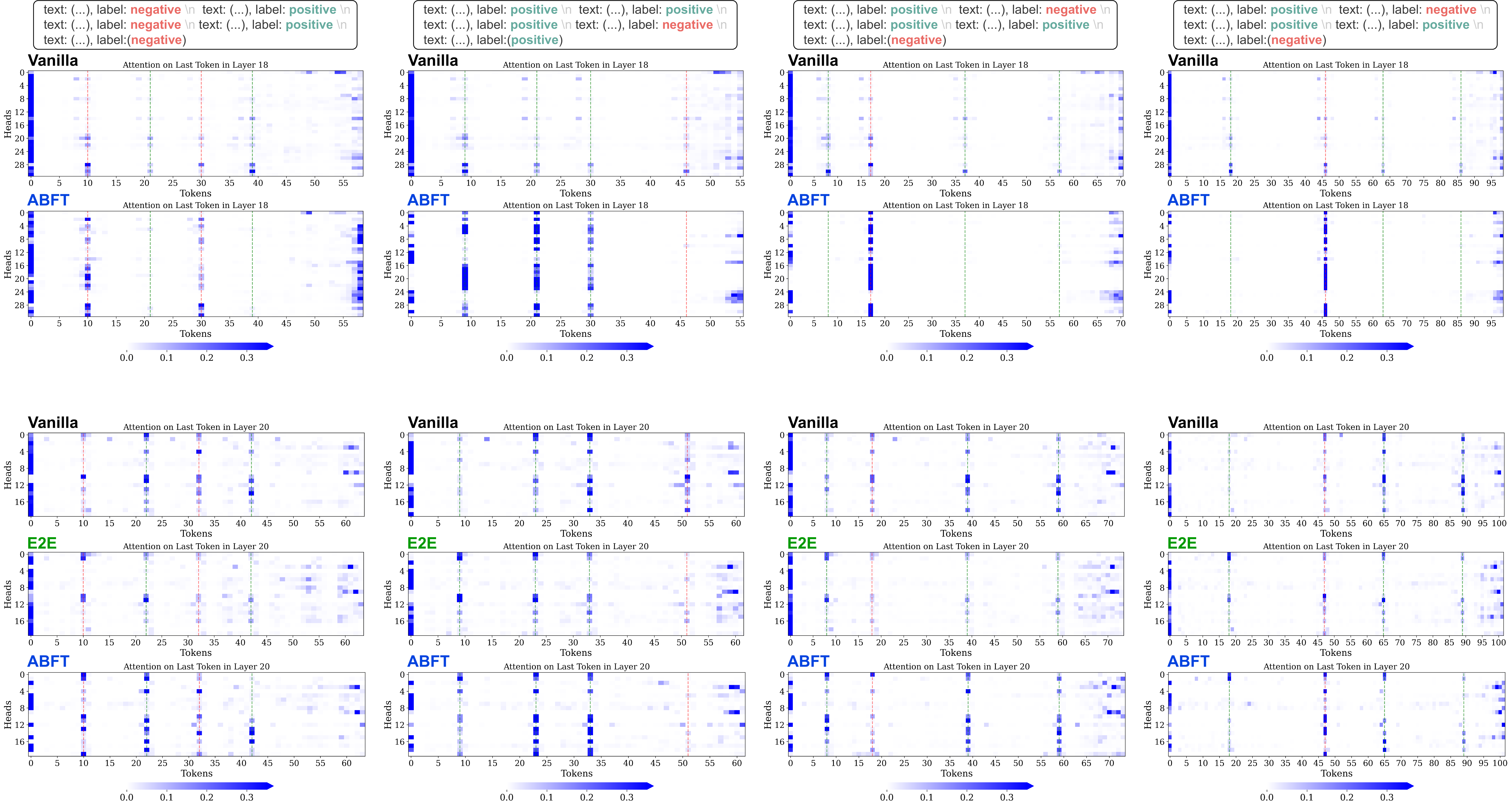}
    \caption{Augmentation results (4 cases) for attention visualization. \textbf{Upper}: results for Fig.~\ref{fig:attention_visual}, \textbf{lower}: results for Fig.~\ref{fig:attention_visualization_3}.}
    \label{appendix.fig:attn}
\end{sidewaysfigure*}

\begin{sidewaysfigure*}[t]
    \centering
    \includegraphics[width=\linewidth]{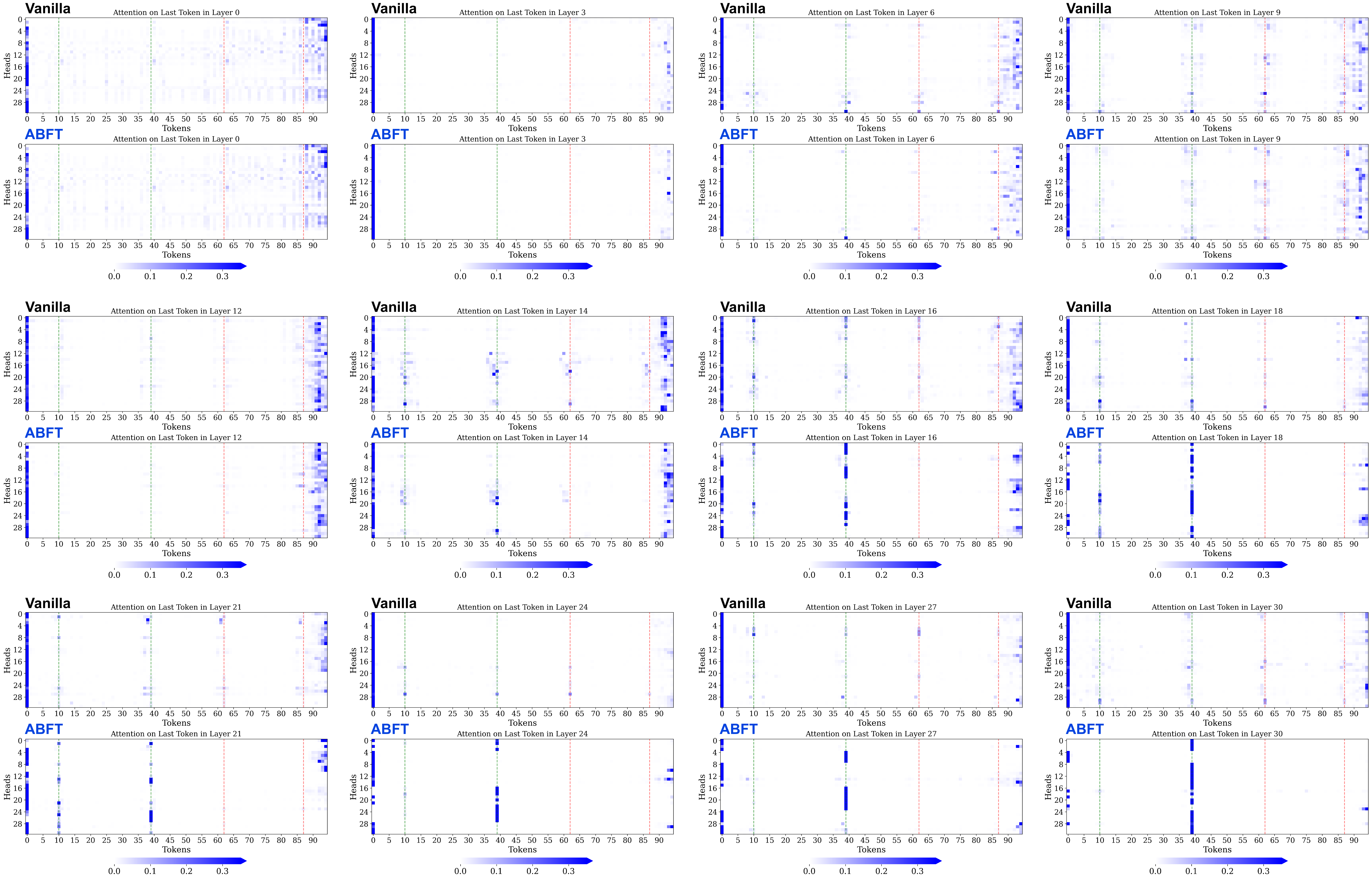}
    \caption{Augmentation results (more layers) for attention visualization with the same settings and input as the Fig.~\ref{fig:attention_visual}.}
    \label{appendix.fig:attn_layers}
\end{sidewaysfigure*}

\begin{sidewaysfigure*}[t]
    \centering
    \includegraphics[width=0.8\linewidth]{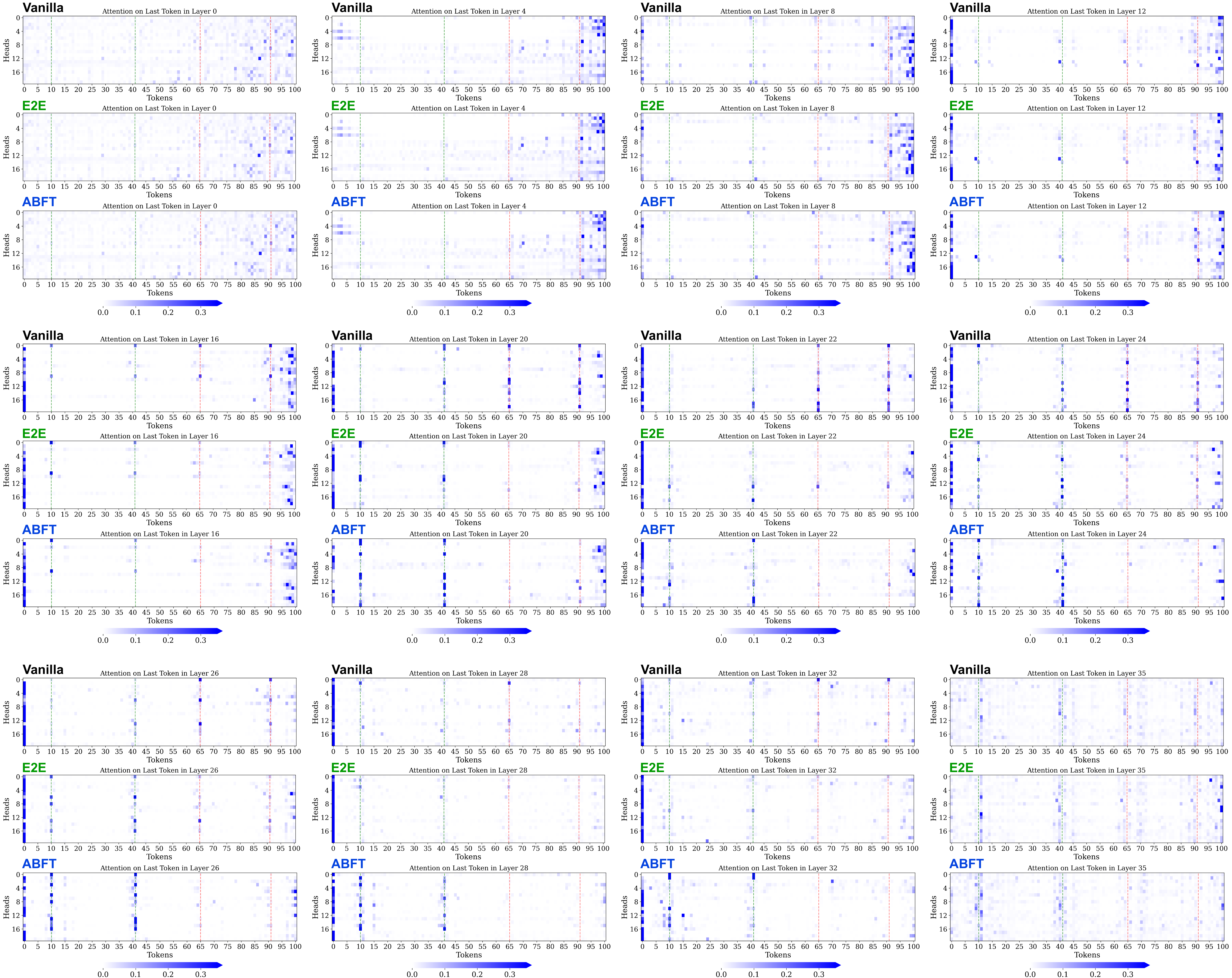}
    \caption{Augmentation results (more layers) for attention visualization with the same settings and input as the Fig.~\ref{fig:attention_visualization_3}.}
    \label{appendix.fig:attn_layers_3}
\end{sidewaysfigure*}

As shown in Fig.~\ref{fig:attention_visual}, we visualize the attention score on the last token of the given input example in the validation set on Llama3 8B, and repeat this visualization on more input cases for Fig.~\ref{fig:attention_visual} and Fig.~\ref{fig:attention_visualization_3} in Fig.~\ref{appendix.fig:attn}. Moreover, we expand the Fig.~\ref{fig:attention_visual} towards more layers in Fig.~\ref{appendix.fig:attn_layers}, and Fig.~\ref{fig:attention_visualization_3} in Fig.~\ref{appendix.fig:attn_layers_3}. We observe that ABFT significantly modifies the attention distribution in the middle layers, while in the early and late layers, neither ABFT nor E2E has a substantial impact on attention scores.

\begin{figure}
    \centering
    \includegraphics[width=1\linewidth]{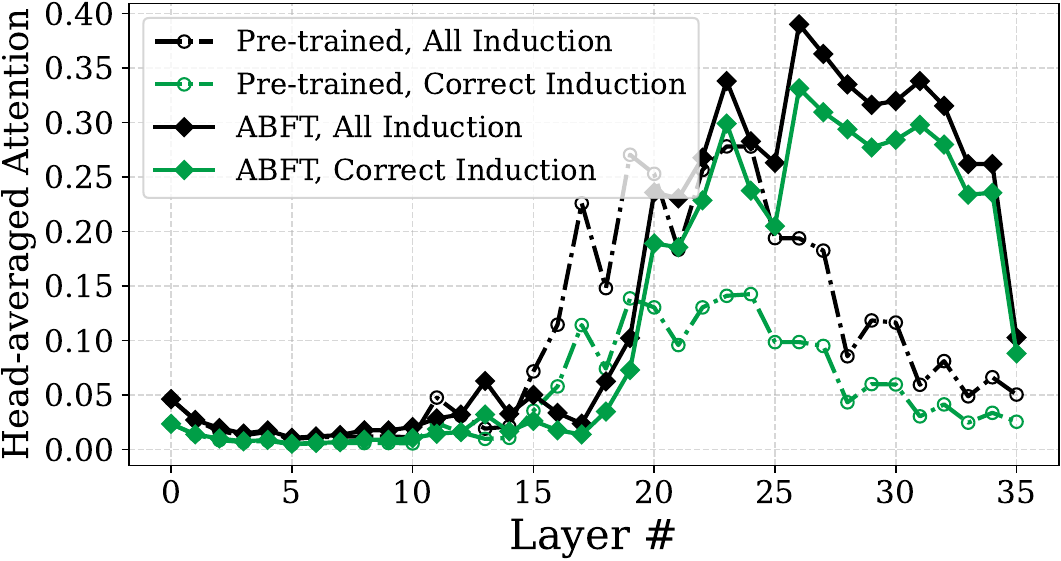}
    \vspace{-1\baselineskip}
    \caption{Augmentation results of Fig.~\ref{fig:attention_score} on GPT2-L.}
    \label{appendix.fig:gpt2_l}
\end{figure}

\begin{figure}
    \centering
    \includegraphics[width=1\linewidth]{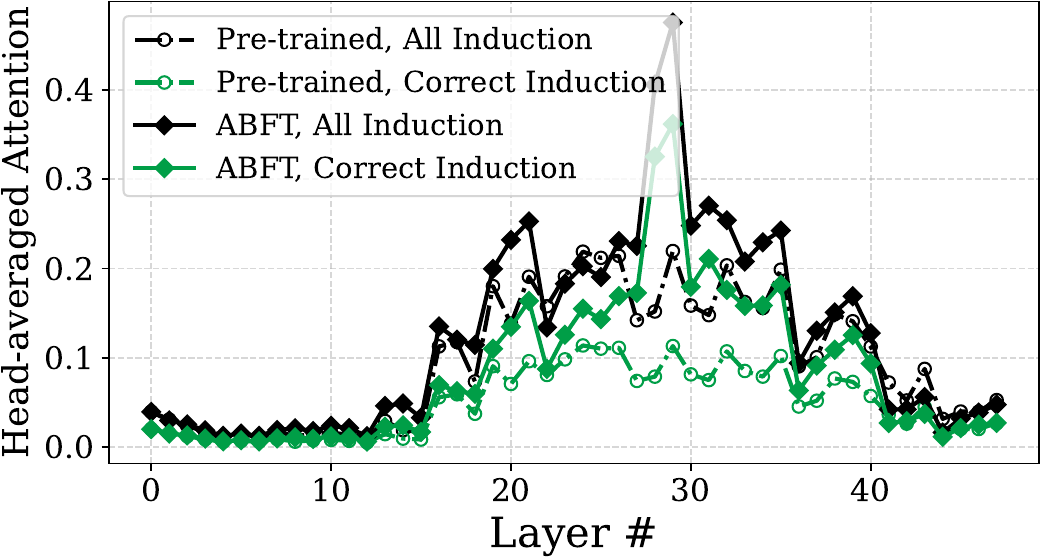}
    \vspace{-1\baselineskip}
    \caption{Augmentation results of Fig.~\ref{fig:attention_score} on GPT2-XL.}
    \label{appendix.fig:gpt2_xl}
\end{figure}

\begin{figure}
    \centering
    \includegraphics[width=1\linewidth]{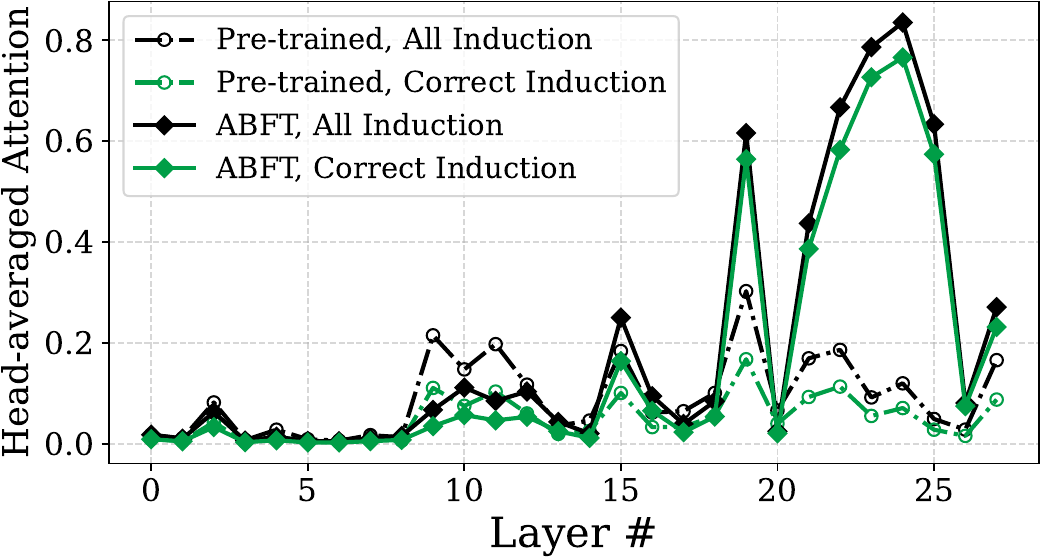}
    \vspace{-1\baselineskip}
    \caption{Augmentation results of Fig.~\ref{fig:attention_score} on Falcon3 7B.}
    \label{appendix.fig:falcon}
\end{figure}

\begin{figure}
    \centering
    \includegraphics[width=1\linewidth]{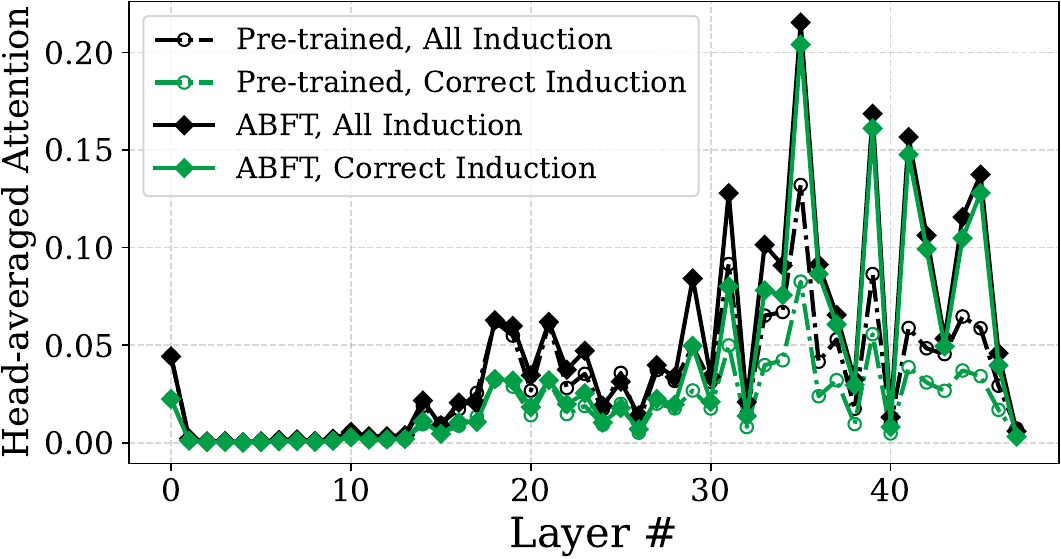}
    \vspace{-0.6\baselineskip}
    \caption{Augmentation results of Fig.~\ref{fig:attention_score} on Llama3 42B.}
    \label{appendix.fig:llama3_42b}
\end{figure}

Moreover, we repeat the attention score visualization similar to Fig.~\ref{fig:attention_score} on more models and SST2, as shown in Fig.~\ref{appendix.fig:gpt2_l},~\ref{appendix.fig:gpt2_xl},~\ref{appendix.fig:falcon}, and~\ref{appendix.fig:llama3_42b}.

\section{Statements}
\label{Appendix:ACS}


\paragraph{Author Contributions Statement.} Hakaze Cho, also known as Yufeng Zhao, handled the entire workload in this paper. He provided ideas, designed / conducted experiments, collected / described data, and wrote / revised the paper. 

Peng Luo suggested the PID algorithm, which is an essential part of this paper, and provided some technical support for it. 

Naoya Inoue, as the supervisor, provided valuable feedback, revisions, and financial support for this paper. 

M.K.\ and R.K.\ made minor contributions that placed them on the borderline of authorship criteria: they participated in discussions and offered comments, some of which were incorporated.

\paragraph{License for Artifacts.} Models and datasets used in this paper are used in their original usage, and are open-sourced with the following license:

\begin{itemize}[topsep=5pt, itemsep=-5pt]
    \item \texttt{mit}: GPT2-L, GPT2-XL, DeepSeek-R1 
    \item Individual License or Unknown: Falcon3, Llama3 8B, Llama3 43B, Llama3 56B, SST5, MR, TREC, SUBJ, TEE, TEH
    \item \texttt{Apache 2.0}: Qwen2.5, SimpleScaling s1.1
    \item \texttt{cc-by-sa-3.0}: SST2, FP, 
\end{itemize}

\paragraph{AI Agent Usage.} AI Agents are used and only used for writing improvement in this paper.

\clearpage

%% file: Table/prompt_templates.tex
\begin{table*}[t] 
    \centering
    \caption{Prompt templates used in this paper.}
    \vspace{-0.5\baselineskip}
    \label{tab:prompt_templates}
    \resizebox{\linewidth}{!}{
    \begin{tabular}{ccc}
    \toprule
      \textbf{Dataset} & \textbf{Prompt Template (Unit)} & \textbf{Label Tokens} \\
    \midrule
      SST2 & \texttt{sentence:\ [input sentence] sentiment:\ [label token] $\backslash$n} & negative, positive \\
      MR & \texttt{review:\ [input sentence] sentiment:\ [label token] $\backslash$n} & negative, positive \\
      FP & \texttt{sentence:\ [input sentence] sentiment:\ [label token] $\backslash$n} & negative, neutral, positive \\
      SST5 & \texttt{sentence:\ [input sentence] sentiment:\ [label token] $\backslash$n} & poor, bad, neutral, good, great \\
      TREC  & \texttt{question:\ [input sentence] target:\ [label token] $\backslash$n} & short, entity, description, person, location, number \\
      SUBJ  & \texttt{review:\ [input sentence] subjectiveness:\ [label token] $\backslash$n} & objective, subjective \\
      TEE & \texttt{tweet:\ [input sentence] emotion: [label token]\ $\backslash$n} & anger, joy, positive, sad \\
      TEH & \texttt{tweet:\ [input sentence] hate speech: [label token]\ $\backslash$n} & normal, hate \\
    \bottomrule
    \end{tabular}
}
\end{table*}

%% file: Table/model_repo.tex
\begin{table}[t]
    \centering
    \caption{\texttt{Huggingface} repository name for models used in this paper.}
    \vspace{-0.5\baselineskip}
    \resizebox{\linewidth}{!}{
    \begin{tabular}{cc}
    \toprule
    \textbf{Model} & \textbf{Repository} \\ \midrule
       GPT2-L & \texttt{openai-community/gpt2-large} \\
       GPT2-XL & \texttt{openai-community/gpt2-xl} \\ 
       Falcon3 & \texttt{tiiuae/Falcon3-7B-Base} \\ 
       Llama3 (8B) & \texttt{meta-llama/Meta-Llama-3-8B} \\
       DeepSeek-R1 & \texttt{deepseek-ai/DeepSeek-R1-Distill-Qwen-14B} \\
       Qwen2.5 & \texttt{Qwen/Qwen2.5-32B} \\
       SimpleScaling s1.1 & \texttt{simplescaling/s1.1-32B} \\
       Llama3 (43B) & \texttt{chargoddard/llama3-42b-v0} \\
       Llama3 (56B) & \texttt{nyunai/nyun-c2-llama3-56B} \\ \bottomrule
    \end{tabular}
    \label{appendix.tab:repo}}
    \vspace{-1\baselineskip}
\end{table}